\documentclass[journal]{IEEEtran}
\usepackage{amsmath,amsfonts}
\usepackage{algorithmic}
\usepackage{algorithm}
\usepackage{array}
\usepackage[caption=false,font=footnotesize,labelfont=rm,textfont=rm]{subfig}
\usepackage{hyperref}
\hypersetup{colorlinks=true,linkcolor=blue,anchorcolor=red,citecolor=blue}
\usepackage{textcomp}
\usepackage{stfloats}
\usepackage{url}
\usepackage{verbatim}
\usepackage{graphicx}
\usepackage{cite}
\usepackage{amssymb}
\usepackage{bbm}
\usepackage{algorithmic}
\usepackage{adjustbox}
\usepackage{multirow}
\usepackage{booktabs}
\usepackage{tabularx}
\begin{document}

\title{A Novel Perception Entropy Metric for Optimizing Vehicle Perception with LiDAR Deployment}

\author{Yongjiang He, Peng Cao, Zhongling Su, and Xiaobo Liu
\thanks{This paper was produced by the IEEE Publication Technology Group. They are in Piscataway, NJ.}
\thanks{This work was supported in part by the National Natural Science Foundation of China under Grant No.52172395. Natural Science Foundation of Sichuan, China No.2022NSFSC0476. (Corresponding author: Xiaobo Liu).}
\thanks{Yongjiang He, Peng Cao, and Xiaobo Liu are with the School of Transportation and Logistics, Southwest Jiaotong University, Chengdu, 610031, China (e-mail: yongjianghe@my.swjtu.edu.cn; caopeng@swjtu.edu.cn; xiaobo.liu@swjtu.cn).}
\thanks{Zhongling Su is with the Shanghai Artificial Intelligence Laboratory, Shanghai, 200232, China (e-mail: suzhongling@pjlab.org.cn).}
}

\markboth{Journal of \LaTeX\ Class Files}%
{Shell \MakeLowercase{\textit{et al.}}: A Sample Article Using IEEEtran.cls for IEEE Journals}


\maketitle
\begin{abstract}
Developing an effective evaluation metric is crucial for accurately and swiftly measuring LiDAR perception performance. One major issue is the lack of metrics that can simultaneously generate fast and accurate evaluations based on either object detection or point cloud data. In this study, we propose a novel LiDAR perception entropy metric based on the probability of vehicle grid occupancy. This metric reflects the influence of point cloud distribution on vehicle detection performance. Based on this, we also introduce a LiDAR deployment optimization model, which is solved using a differential evolution-based particle swarm optimization algorithm. A comparative experiment demonstrated that the proposed PE-VGOP offers a correlation of more than 0.98 with vehicle detection ground truth in evaluating LiDAR perception performance. Furthermore, compared to the base deployment, field experiments indicate that the proposed optimization model can significantly enhance the perception capabilities of various types of LiDARs, including RS-16, RS-32, and RS-80. Notably, it achieves a 25\% increase in detection \textit{Recall} for the RS-32 LiDAR.
\end{abstract}

\begin{IEEEkeywords}
LiDAR deployment; evaluation metric; vehicle perception; perception entropy; optimization method
\end{IEEEkeywords}

\section{Introduction}
\IEEEPARstart{L}{IGHT} Detection and Ranging (LiDAR) can generate detailed 3D spatial data in real-time, based on which traffic objects can be detected and tracked \cite{10208208,zhou2022leveraging,wu2020automatic}. To advance the progress of connected and automated vehicles (CAVs), the deployment of LiDARs on vehicles or at roadside units is becoming increasingly prevalent for gathering detailed traffic information. However, differences in detection accuracy among various LiDAR deployments range from 5\% to 10\%, which can cause significant difficulties in motion planning, vehicle control, obstacle avoidance, and other autonomous driving functions \cite{cai2023analyzing,xu2023opencda,hu2022investigating}. Therefore, strategically deploying LiDAR systems to maximize their perception capabilities is imperative. In the quest to identify the optimal LiDAR deployment, evaluating the perception performance of different deployments is essential. Consequently, developing an accurate and efficient evaluation metric for LiDAR perception performance is crucial. 

Current research on LiDAR perception performance evaluation metrics can be divided into two categories. The first category is object detection-based metrics, which evaluate LiDAR perception performance based on object detection algorithms \cite{roos2021framework,xu2021spg}. Metrics such as \textit{Recall}, mean Average Precision (\textit{mAP}), Intersection over Union (\textit{IoU}), and \textit{confidence} are widely used to evaluate LiDAR performance. These metrics are considered the ground truth for evaluating LiDAR perception performance because they accurately reflect LiDAR's capabilities in vehicle detection, classification, and localization \cite{chen2017multi,wu2020deep,mao20223d}. However, object detection-based metrics rely on complex deep learning and clustering algorithms, which require time-consuming training and significant computing power. The second category is point cloud data-based metrics, which encompass parameters such as point cloud coverage area, numbers, and density \cite{lambert2020performance,cai2023analyzing,li2024optimizing,arefkhani2023sensor}. These metrics are derived through straightforward statistical analyses, demanding minimal computational resources to offer a rapid assessment of LiDAR perception performance. However, existing point cloud data-based metrics overlook the impact of vehicle point cloud distribution on perception performance, resulting in an inability to accurately evaluate LiDAR perception performance. 

Optimizing LiDAR deployment in real-world scenarios is challenging. One major issue is the lack of metrics that can simultaneously generate fast and accurate evaluations based on either object detection or point cloud data. This makes it difficult to effectively assess LiDAR perception performance \cite{10413581}. Additionally, to enhance LiDAR perception capability, some studies have proposed optimal deployment configurations through simulation tools and optimization methods \cite{moshiri2017evaluation,cao2019adversarial}. However, existing simulation tools are limited to simulating LiDARs with uniformly distributed beams and cannot handle those with uneven beam distribution \cite{hu2022investigating}. Therefore, it is necessary to develop a simulator capable of simulating various types of LiDAR, configuring deployments, and generating point clouds. Finally, previous optimization methods mainly focus on the number and placement of LiDARs, with limited studies addressing the tilt angle of LiDAR deployment \cite{9055238}. The tilt angle alters the direction of LiDAR beams, influencing the distribution of point clouds on detected vehicles \cite{ji2023optimization}. To maximize LiDAR perception capability, it is essential to develop an optimization model for LiDAR deployment that includes both placement and tilt angle. 

\begin{figure}[!ht]
\centering
\includegraphics[width=3.0in]{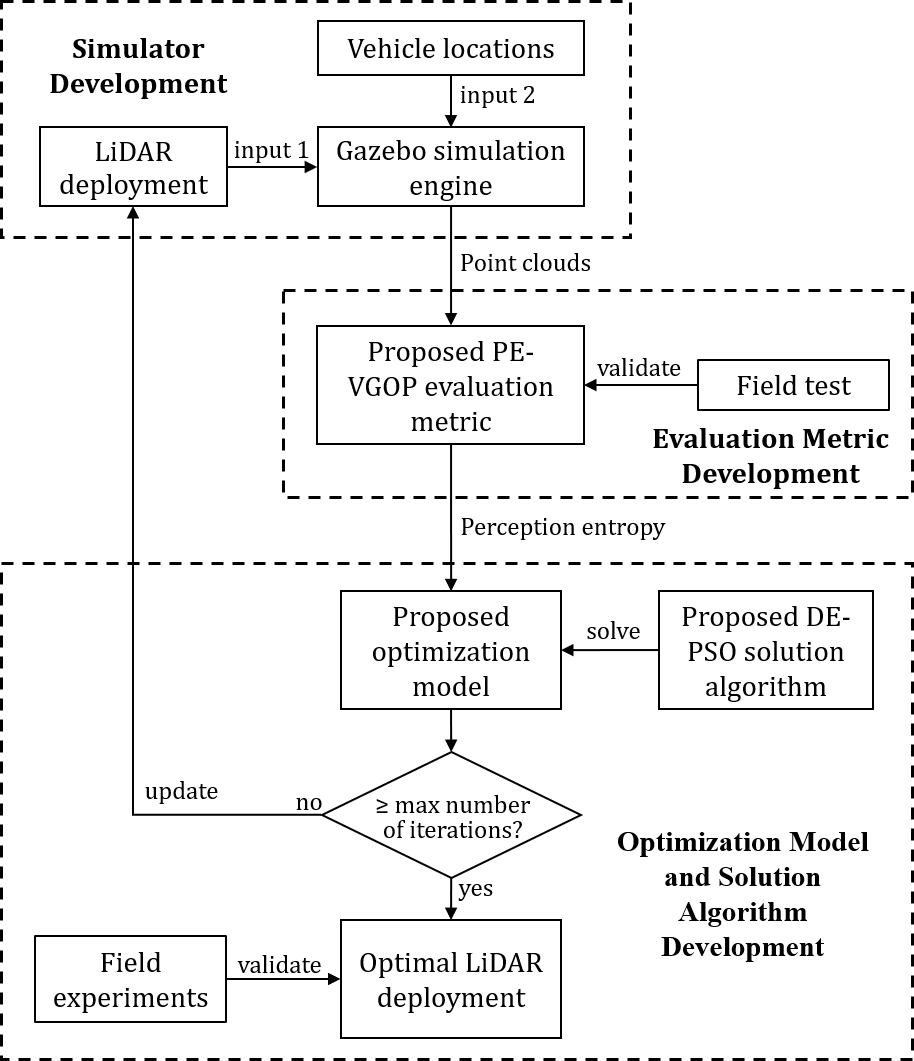}
\caption{LiDAR deployment optimization framework.}
\label{fig1}
\end{figure}

To address these three challenges, this paper proposes a LiDAR deployment optimization framework, as shown in Fig.~\ref{fig1}. The main contributions of this study are summarized as follows:

\begin{enumerate}
  \item We proposed a novel metric called Perception Entropy based on Vehicle Grid Occupancy Probability (PE-VGOP). This metric leverages the distribution of vehicle point clouds to swiftly and accurately assess vehicle detection performance under different LiDAR deployments.
  \item We developed a LiDAR deployment simulator based on Gazebo. The simulator can model various types of LiDAR, customize LiDAR deployments, simulate traffic scenarios, and generate point clouds.  
  \item We developed a LiDAR deployment optimization model focused on placement and tilt angle with the goal of maximizing perception entropy. To solve the proposed optimization model, we designed a differential evolution-based particle swarm optimization algorithm.  
\end{enumerate}

\section{Related Works}
\subsection{Point Cloud-based LiDAR Performance Evaluation}
Previous research on point cloud data-based LiDAR performance evaluation metrics has primarily focused on statistically analyzing the inherent physical properties of point clouds, such as quantity, density, and coverage area \cite{ma2021perception,liu2021survey,zhou2019iou}. For example, Vijay \textit{et al}. \cite{vijay2021optimal} divided the road surface into grids and evaluated LiDAR perception performance at different deployments by counting the number of covered grids. Jin \textit{et al}. \cite{jin2022novel} used the density of point clouds in the Region of Interest (RoI) to evaluate the LiDAR perception performance at various heights. Roos \textit{et al}. \cite{roos2021framework} proposed dividing the 3D space occupied by vehicles into voxels and then converting the proportion of occupied voxels into perception entropy to evaluate vehicle detection performance. Similarly, Hu \textit{et al}. \cite{hu2022investigating} suggested converting the voxel occupancy probability of point clouds in the ROI into information gain, thus assessing LiDAR perception capability on CAVs. However, existing point cloud data-based metrics do not account for the impact of point cloud distributions on detection results, leading to inaccurate evaluations of LiDAR perception capabilities in many scenarios \cite{liu2021survey,ma2021perception}. For instance, in heavy traffic environments, these metrics might indicate very high LiDAR perception performance due to the large number of point clouds. However, these evaluations are misleading because the point clouds may be concentrated on a limited number of vehicles, causing many other vehicles to remain undetected due to sparse LiDAR points. Therefore, this study aims to propose an evaluation metric that considers the point cloud distributions on vehicles, providing a more accurate assessment of LiDAR perception performance.

\subsection{LiDAR Deployment Optimization}
Existing studies on LiDAR deployment optimization have mainly focused on the number and placement of LiDARs \cite{liu2019should,mou2018optimal,qu2023seip}. For example, Jiang \textit{et al}. \cite{jiang2023optimizing} proposed a greedy algorithm based on perceptual gain to optimize the placement of roadside multi-LiDARs and achieve optimal perception; Kim \textit{et al}. \cite{kim2019placement} developed a genetic algorithm to optimize placement of LiDAR on a vehicle, aiming to reduce dead zones and improve point cloud resolution; Ji \textit{et al}. \cite{ji2023optimization} proposed an optimization model for the positional relationship and point cloud coverage of roadside multi-LiDARs to optimize their placement on urban roads. Only a few works have focused on the tilt angle in LiDAR deployment optimization. For example, Mou \textit{et al}. \cite{mou2018optimal} proposed an optimization model that maximizes grid occupancy in the perception area to optimize LiDAR placement and tilt angle on CAVs. However, including the tilt angle as a parameter in LiDAR deployment dramatically increases the complexity and solving difficulty of the optimization model, because the tilt angle alters the original emission directions and perception space of all LiDAR beams  \cite{ma2021perception}. For instance, Jin \textit{et al}. \cite{jin2022novel} optimized a roadside LiDAR deployment by comparing the perception performance at different installation heights and tilt angles using a traversal approach due to the lack of effective optimization models and solution algorithms. These gaps in models and algorithms result in limited perception capabilities of LiDAR in real-world applications. Therefore, this study aims to propose a LiDAR placement and tilt angle deployment optimization model and design an efficient algorithm to maximize LiDAR's vehicle perception capability.

\section{Methodologies}
\subsection{Development of LiDAR Deployment Simulator}
This study focuses on mechanical rotating LiDARs, defining the vertical angles of its beams as $\beta$. Each laser beam, denoted as $\beta_i \in \beta$, generates varying horizontal angles ($\alpha_i \in \alpha$) by horizontally rotating the beam at a specified resolution. Thus, a LiDAR can be formalized as a collection of spatial vectors represented by $\beta$ and $\alpha$:
\begin{equation}
\label{equ-3-1}
L(\alpha,\beta)  = \begin{bmatrix}\sin\left (\alpha   \right ) \cos \left ( \beta  \right ) 
 \\\cos\left (\alpha   \right ) \cos \left ( \beta \right ) 
 \\ \sin \left ( \beta \right ) 
\end{bmatrix}
\end{equation}

Eq. (\ref{equ-3-1}) represents the perceptual space of the LiDAR. The perception space will change with LiDAR deployments and can be formalized as follows:
\begin{equation}
\label{equ-3-2}
L'(\alpha,\beta) = L(\alpha,\beta) \cdot R_{tilt} +L_{lidar}
\end{equation}
where $L_{lidar}$ represents the location of the LiDAR, and $R_{tilt}$ represents the tilt matrix of the LiDAR. The tilt matrix for rotating around the X and Y axes by a tilt angle $\theta$ is given by:

\begin{equation}
\label{equ-3-2-a}
R_x(\theta) = \begin{bmatrix}
1 & 0 & 0 \\
0 & \cos(\theta) & -\sin(\theta) \\
0 & \sin(\theta) & \cos(\theta)
\end{bmatrix}
\end{equation}

\begin{equation}
\label{equ-3-2-b}
R_y(\theta) = \begin{bmatrix}
\cos(\theta) & 0 & \sin(\theta) \\
0 & 1 & 0 \\
-\sin(\theta) & 0 & \cos(\theta)
\end{bmatrix}
\end{equation}

\begin{equation}
\label{equ-3-2-c}
R_{tilt} = R_x(\theta_x) \cdot R_y(\theta_y)
\end{equation}

The tilt matrix for Z-axis rotation does not need to be considered since the rotation of the LiDAR around the Z-axis does not affect the emission angles of the laser beams. Therefore, the point cloud collected by LiDAR in a specific deployment configuration can be represented as:

\begin{equation}
\label{equ-3-3}
Points = d \times L'(\alpha,\beta)
\end{equation}
where $d$ represents the propagation distance of the lasers from the LiDAR unit to the target.

To collect point cloud data across diverse LiDAR deployments, we developed a simulator by leveraging the secondary development capabilities of the Gazebo sensor model \cite{koenig2004design}. As illustrated in Fig.~\ref{fig2}, the steps for point cloud data collection using the simulator are as follows:

\begin{figure}[!ht]
\centering
\includegraphics[width=3.4in]{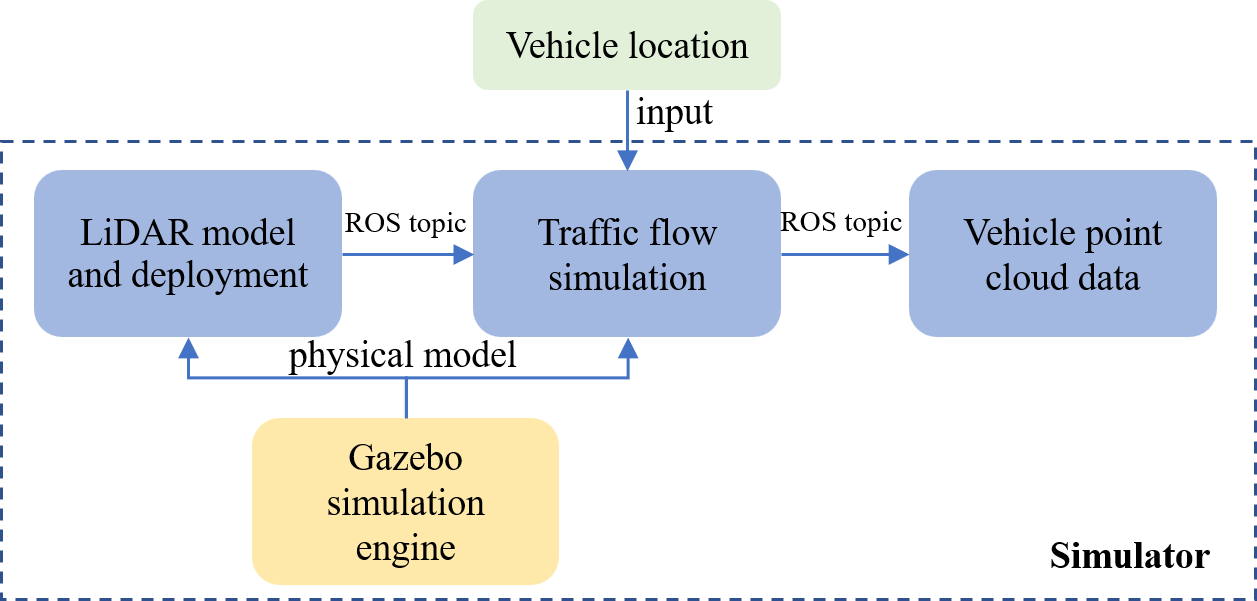}
\caption{Simulator for LiDAR deployment and vehicle point cloud collection.}
\label{fig2}
\end{figure}

\textit{Step 1}: Define a LiDAR model as in Eq.~(\ref{equ-3-1}) and set the deployment configurations as Eq.~(\ref{equ-3-2}) - (\ref{equ-3-2-c}).

\textit{Step 2}: Obtain vehicle locations from traffic simulation software or trajectory datasets, and generate corresponding vehicle models in the Gazebo simulator.

\textit{Step 3}: Publish laser beams at various angles as \textit{ROS topic} messages.

\textit{Step 4}: Utilize Gazebo’s physical sensor model for ray tracing and collision detection to determine the intersection points of laser beams within the vehicle.

\textit{Step 5}: Publish the intersection points as \textit{ROS topic} messages and save them as vehicle point clouds.

The developed simulator can customize various LiDAR models and deployment configurations, and collect vehicle point clouds in different scenarios. By evaluating the point clouds collected under different deployment configurations, LiDAR deployment optimization can be achieved.

\subsection{Modeling of the PE-VGOP}

In this section, a novel metric named Perception Entropy based on Vehicle Grid Occupancy Probability (PE-VGOP) is introduced to assess LiDAR perception performance using the collected point clouds. The modeling process is detailed as follows:

Initially, the collected point clouds are transformed from the LiDAR coordinate system to the vehicle coordinate system to characterize their distributions on the vehicle. As shown in Fig.~\ref{fig3}, the vehicle coordinate system is established as a right-hand system, with the vehicle's forward direction serving as the X-axis and the center of the vehicle's bounding box as the origin. The transformation process can be formalized as follows:

\begin{figure}[!ht]
\centering
\includegraphics[width=3.4in]{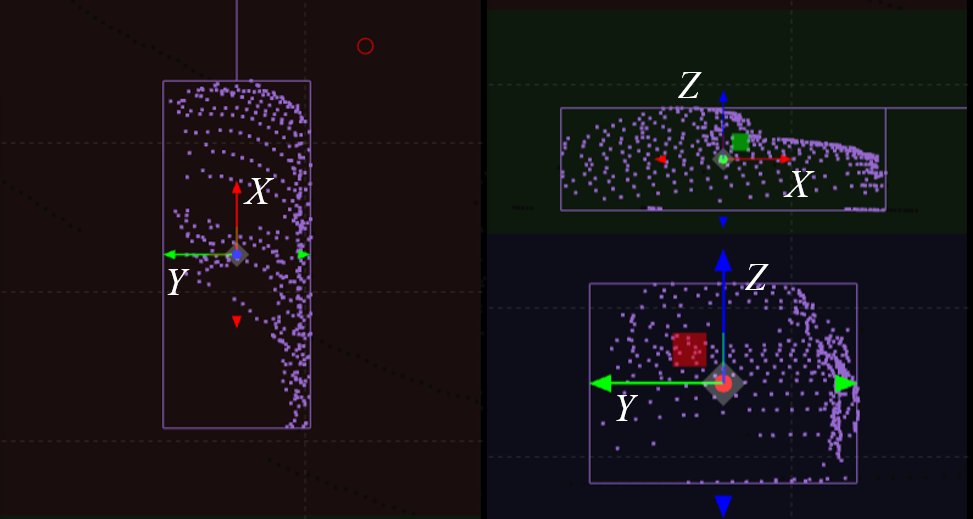}
\caption{Visualization of vehicle coordinate system.}
\label{fig3}
\end{figure}

\begin{equation}
\label{equ-3-4}
P_{veh} = \left ( P_{lidar} - C_{box}  \right ) \cdot R_{yaw}
\end{equation}
where $P_{veh}$ and $P_{lidar}$ represent point cloud coordinates in the vehicle and LiDAR coordinate systems, respectively. $C_{box}$ represents the center coordinate of the vehicle's bounding box in the LiDAR coordinate system. $R_{yaw}$ denotes the vehicle's rotation matrix, which is calculated as:
\begin{equation}
\label{equ-3-5}
R_{yaw} = \begin{bmatrix} \cos \left ( \theta_v \right )   & -\sin \left ( \theta_v \right )  & 0\\ \sin \left ( \theta_v \right )  & \cos \left ( \theta_v \right ) & 0 \\ 0  & 0 & 1\end{bmatrix}
\end{equation}
where $\theta_v$ represents the vehicle's heading angle. 

Subsequently, vehicle point clouds are projected from three-dimensional space onto three two-dimensional planes along the X, Y, and Z-axes of the vehicle coordinate system to obtain three orthographic projection perspectives: front view, side view, and top view. Each projection plane is divided into identical grids. The state of each grid, either occupied or unoccupied by point clouds, is represented as $v_i$ and is calculated as follows:

\begin{equation}
\label{equ-3-6}
v_i = 
\begin{cases} 
1, & \text{if } \exists p_i \in g_i \\
0, & \text{if } \nexists  p_i \in g_i 
\end{cases}
\end{equation}
where $g_i$ represents the grid $i$ in a projection plane and $p_j=(x_j,y_j,z_j)$ represents a point on the projection plane.

\begin{figure}[!ht]
\centering
\subfloat[top-down view]{
\includegraphics[width=2.0in]{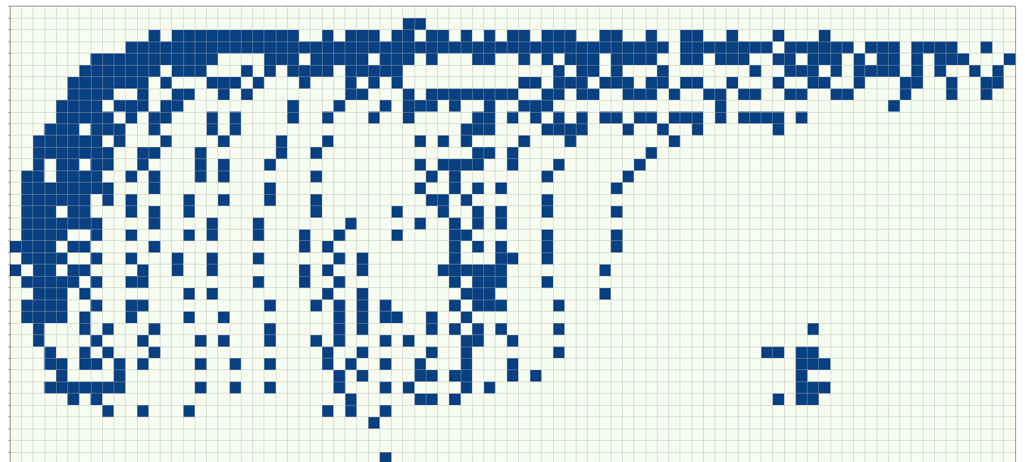}}
\hfill
\subfloat[front view]{
\includegraphics[width=1.3in]{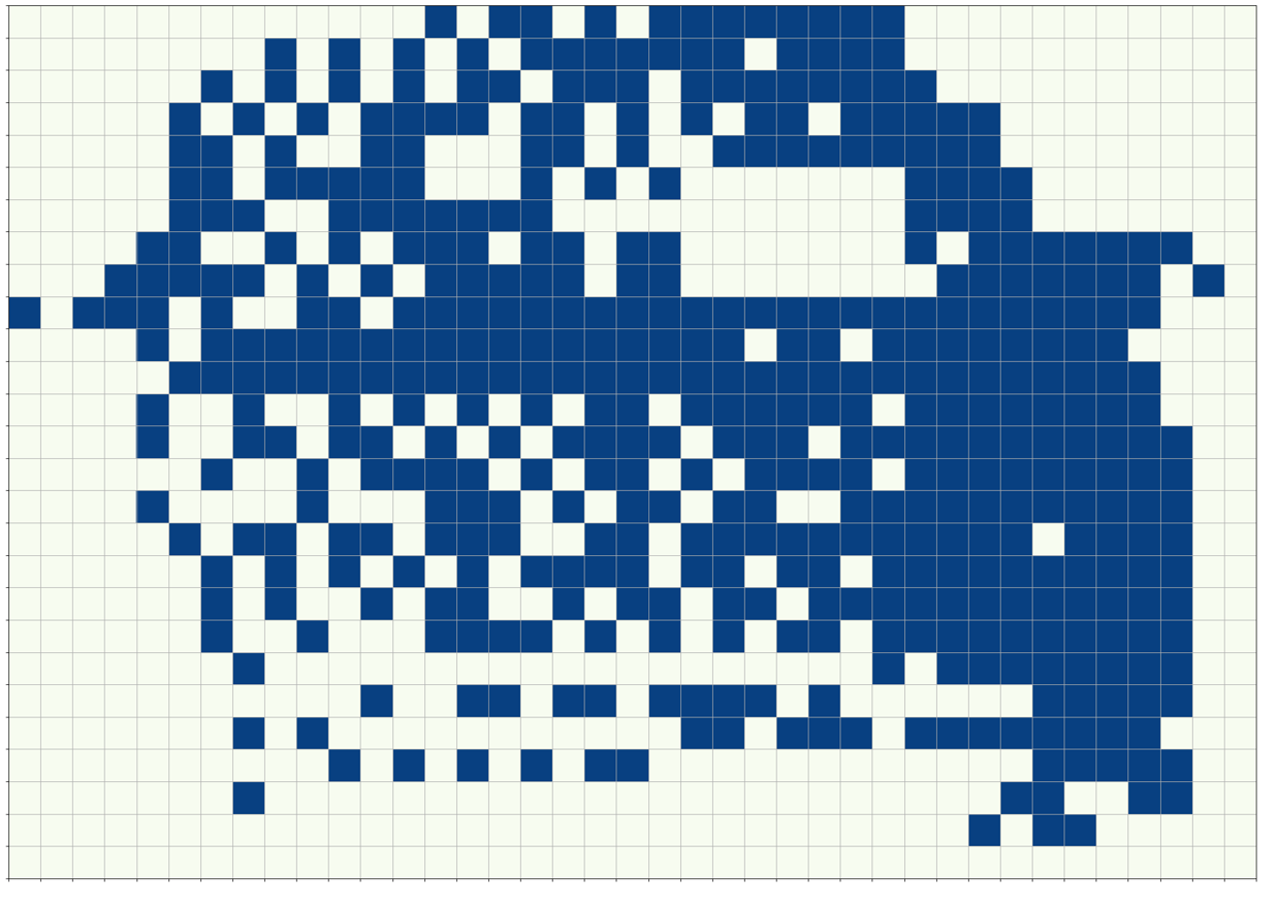}}
\hfill
\subfloat[side view]{
\includegraphics[width=3.4in]{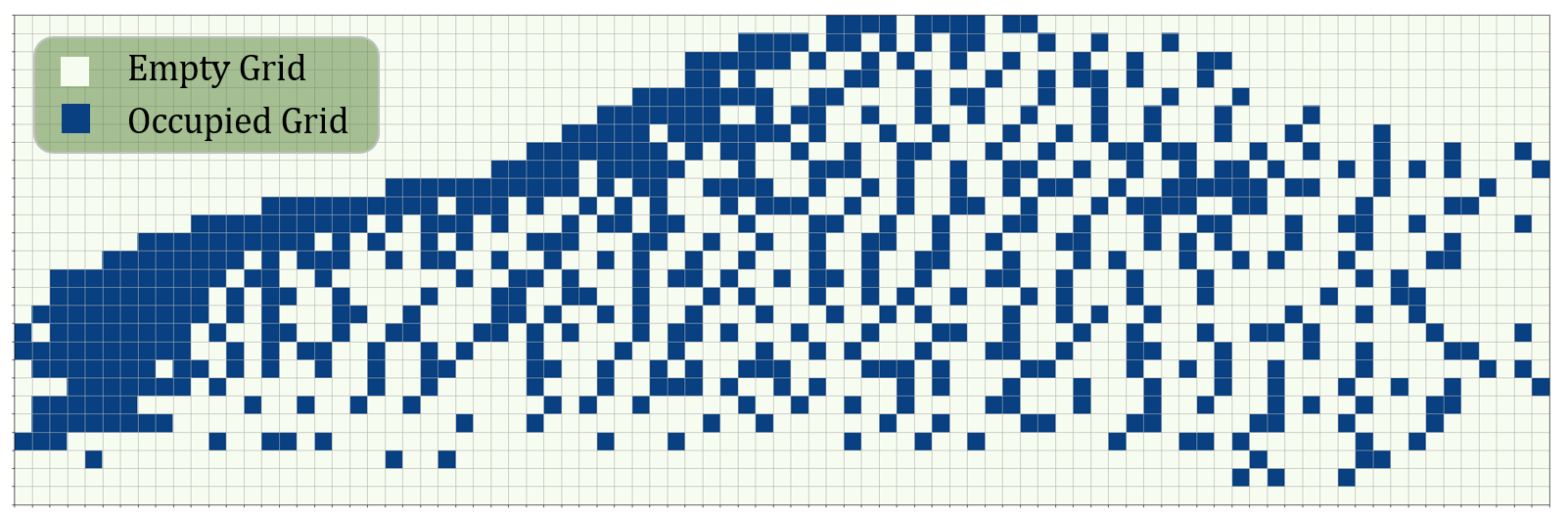}}
\caption{Example of VGOP with the grid dimensions $\mu_t=\mu_s=\mu_f=0.05 \text{m} \times 0.05 \text{m}$.}
\label{fig4}
\end{figure}

Additionally, the VGOP in each projection plane is used to represent the quantity, density, and distribution characteristics of the vehicle point clouds, as shown in Fig.~\ref{fig4}. These characteristics can be derived through statistical calculations as:
\begin{equation}
\label{equ-3-7}
\left\{\begin{matrix}P \left ( v^{t}  \right ) =\frac{\sum\limits_{i=1}^{N_t}\left ( v_{i}^{t}  \right )}{N_t}   \\P \left ( v^{s}  \right ) = \frac{\sum\limits_{i=1}^{N_s}\left ( v_{i}^{s}  \right )}{N_s}   \\P \left ( v^{f}  \right ) = \frac{\sum\limits_{i=1}^{N_f}\left ( v_{i}^{f}  \right )}{N_f} \end{matrix}\right.
\end{equation}
where $P(v^{t})$, $P(v^{s})$, and $P(v^{f})$ represent the VGOP under top view $t$, side view $s$, and front view $f$, respectively. $N_t$, $N_s$, and $N_f$ represent the number of grids in top view, side view, and front view and are calculated as follows:
\begin{equation}
\label{equ-3-8}
N_t=\frac{l \times w}{\mu_t}, N_s=\frac{l \times h}{\mu_s} ,N_f=\frac{w \times h}{\mu_f} 
\end{equation}
where $\mu_t$, $\mu_s$, and $\mu_f$ represent the grid dimensions in the top view, side view, and front view, respectively. $l$, $w$, and $h$ represent the length, width, and height of the vehicle, respectively.

Finally, the PE-VGOP is employed to quantify the amount of information contained within vehicle point clouds, which is calculated as:
\begin{equation}
\label{equ-3-9}
E(veh) = - \sum\limits_{j \in\{t,s,f\}}P \left ( v^{j}  \right ) \log_{2}{\left ( P \left ( v^{j}  \right )  \right)}
\end{equation}
where $E(v_i)$ represents the PE-VGOP of a vehicle. As a alternative metric for evaluating vehicle detection performance, the effectiveness of PE-VGOP is demonstrated in Section IV.

\subsection{LiDAR Deployment Optimization Model}

The optimization objective is defined as maximizing the perception entropy of all vehicles in the perception area with the LiDAR deployment of location $L_{lidar}$ and tilt matrix $R_{tilt}$, which is formulated as:

\begin{equation}
\label{equ-3-10}
\begin{aligned}
F(L_{lidar},R_{tilt})^* &= \arg\max \sum_{i=1}^{N}  E(veh_i) \mathbbm{1}(P(v_i)) \\
&\quad - C (1-\mathbbm{1}(P(v_i)))  
\end{aligned}
\end{equation}

\begin{equation}
\label{equ-3-11}
\mathbbm{1}(P(v_i)) = 
\begin{cases} 
1, & \text{if } P(v_i) \ge \delta \\
0, & \text{if } P(v_i)  < \delta 
\end{cases}
\end{equation}
where $\mathbbm{1}(P(v_i))$ represents the indicator function, which takes the value of one when $P(v_i) \ge \delta$ and 0 otherwise; $C$ represents the constant loss; $\delta$ represents the threshold of VGOP, indicating that a vehicle can be detected when the VGOP exceeds a certain threshold; otherwise, it may result in detection loss.

Considering the non-linearity and non-convexity of the optimization model, we developed a differential evolution-based particle swarm optimization algorithm (DE-PSO) to solve the model. The algorithm procedure is shown in Algorithm 1, and the main steps are detailed as follows:

\begin{algorithm}[!ht]
\label{alg-1}
\caption{DE-PSO algorithm}
\begin{algorithmic}[1]
\renewcommand{\algorithmicrequire}{\textbf{Input:}} 
\REQUIRE ~~\\
Iteration number $T$; particle swarm number $N$; adaptation function $F(L_{lidar},R_{tilt})$; inertia weighting $w_1$; differential weighting $w_2$; acceleration factors $a_1$, $a_2$; differential threshold $\gamma$ 
\STATE \textbf{Initialize:}
Positions of particle swarms $P_i=(L_i,R_i)$; velocities $V_i=0$; personal best positions $P_i^{best}$; global best position $G_{best}$
\FOR{$t=1$ to $T$}
    \FOR{$i=1$ to $N$}
        \STATE $r_1$, $r_2$ $\gets$ random numbers between 0 and 1
        \STATE update velocity of $P_i$ use Eq. (\ref{equ-3-12})   
        \STATE $r_3$ $\gets$ random number between 0 and 1
        \IF{$r_3$ $< \gamma$}
            \STATE apply differential evolution strategy use Eq. (\ref{equ-3-14})
        \ENDIF
        \STATE update position of $P_i$ use Eq. (\ref{equ-3-13})
        \STATE apply constraints to position use Eq. (\ref{equ-3-15})
        \STATE calculate $F(L_i,R_i)$ use Eq. (\ref{equ-3-10})
        \IF{$F(L_i,R_i) > P_i^{best}$}
            \STATE  $P_i^{best}=(L_i,R_i)$
        \ENDIF
        \IF{$F(L_i,R_i) > G_{best}$}
            \STATE  $G_{best}=(L_i,R_i)$;
        \ENDIF
    \ENDFOR
\ENDFOR
\renewcommand{\algorithmicensure}{\textbf{Output:}} 
\ENSURE ~~\\%
Optimal solution $(L_{lidar},R_{tilt})^*$
\end{algorithmic}
\end{algorithm}

\subsubsection{Update Velocities and Positions of Particle Swarms}
\
\par
During the iteration process, each particle moves towards its historical personal best position and the global best position, influenced by inertia weight and acceleration factors. The movement speed is calculated as:
\begin{equation}
\label{equ-3-12}
\begin{aligned}
V_{i}(t+1) = & w_1 \cdot V_{i}(t) + a_1 r_{1} \cdot (P_i^{best}- P_i(t)) \\
& + a_2 r_{2} \cdot (G_{best}- P_i(t))
\end{aligned}
\end{equation}
where $V_{i}(t+1)$ and $V_{i}(t)$ represent the velocity of particle $i$ at time $t+1$ and $t$, respectively; $w_1$ is the inertia weight; $a_1$ and $a_2$ are the cognitive and social coefficients, respectively; $r_1$ and $r_2$ are random numbers between 0 and 1; $P_i^{best}$ is the personal best position of particle $i$; $P_i(t)$ is the current position of particle $i$ at time $t$; $G_{best}$ is the global best position among all particles.

Based on the velocity, the position of the particle swarm is updated as:
\begin{equation}
\label{equ-3-13}
\begin{aligned}
P_{i}(t+1) = P_{i}(t) + V_{i}(t+1)
\end{aligned}
\end{equation}
where $P_{i}(t+1)$ represents the position of the particle at time $t+1$.

\subsubsection{Differential Evolution Strategy}
\
\par
During each iteration, a random number $r_3$ between 0 and 1 is assigned to each particle. If this number falls below the differential threshold $\gamma$, the differential evolution strategy is implemented for that particle.
\begin{equation}
\label{equ-3-14}
\begin{aligned}
V_{i}(t+1) = P_{i}(t) + w_2  \cdot (P_j(t) - P_k(t))
\end{aligned}
\end{equation}
where $w_2$ represents the differential weight; $P_j$ and $P_k$ represent the positions of two randomly selected particles $j$ and $k$, respectively.

\subsubsection{Constraints Strategy}
\
\par
The random parameters and differential evolution strategy may lead to particles' positions exceeding the constraint space. Eq. (\ref{equ-3-15}) addresses this issue by constraining the positions of particles that exceed the boundaries to their respective boundaries.
\begin{equation}
\label{equ-3-15}
P_{i,j}(t+1) = \begin{cases} 
P_j^-, & \text{if } P_{ij}(t+1) < P_j^- \\
P_j^+, & \text{if } P_{ij}(t+1) > P_j^+ \\
P_{ij}(t+1), & \text{otherwise}
\end{cases}
\end{equation}
where $P_{i,j}$ represents the position of particle $i$ in dimension $j$; $P_j^-$ and $P_j^+$ represent the lower and upper bounds of dimension $j$, respectively.

\section{Validate of PE-VGOP}
This section validates the proposed PE-VGOP metric for evaluating LiDAR perception performance. In support of this proposal, the proposed PE-VGOP was utilized to assess the detection performance of 14,357 vehicles at different locations within 3,712 selected point cloud frames from the KITTI dataset \cite{Geiger2012CVPR}. These selected frames feature real-world point cloud data from a Velodyne 64-beam LiDAR, capturing diverse environments such as urban and rural areas. Additionally, we compared the PE-VGOP with state-of-the-art (SOTA) metrics, including MDG-P \cite{jin2022novel}, S-MIG \cite{hu2022investigating}, and perception entropy \cite{roos2021framework}. In the comparison experiment, the vehicle detection performance based on the PointPillars vehicle detection algorithm \cite{lang2019pointpillars} was used as the ground truth, which can be calculated as:

\begin{equation}
\label{equ-4-1}
Performance = conf \times IoU
\end{equation}
where $conf$ denotes the confidence level that a vehicle is within the predicted bounding box, and $IoU$ denotes the intersection over union between the predicted and ground truth bounding boxes. The $Performance$ metric integrates prediction accuracy, as measured by $IoU$, with prediction confidence, as measured by $conf$.

\begin{figure}[!ht]
\centering
\includegraphics[width=3.4in]{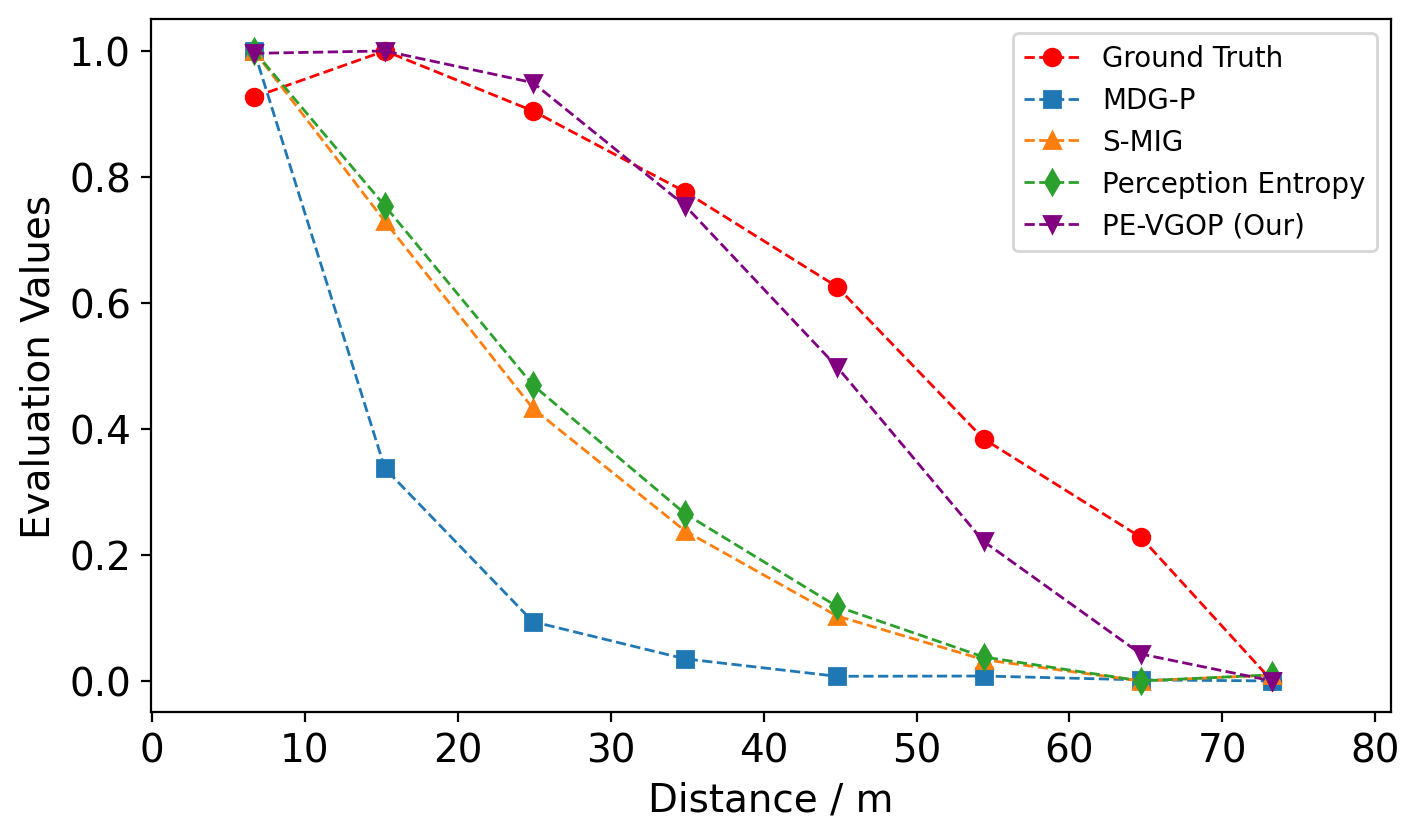}
\caption{Comparison of the proposed PE-VGOP and SOTA metrics in evaluating LiDAR perception performance.}
\label{fig5}
\end{figure}

Fig.~\ref{fig5} illustrates the evaluation results of LiDAR perception performance using alternative metrics and the ground truth of the vehicle detection algorithm.  Although it is evident that all alternative metrics capture the general trend of diminishing LiDAR perception performance with increasing distance, there are discrepancies in the finer details. Specifically, the evaluation results of SOTA metrics indicate that the LiDAR perception capability for vehicles follows a logarithmic decrease with increasing distance. However, the proposed PE-VGOP metric demonstrates a linear decrease in LiDAR perception capability, more closely aligning with the ground truth of the vehicle detection algorithm. This result indicates that the proposed PE-VGOP reflects LiDAR perception capabilities more accurately compared to SOTA metrics. More importantly, using the PE-VGOP to evaluate 3,712 frames of data takes only 18 seconds, which is significantly faster than the evaluation efficiency of the vehicle detection algorithm.

\begin{figure*}[!htt]
\centering
\subfloat[MDG-P \cite{jin2022novel}]{\label{fig6a}
\includegraphics[width=2.2in]{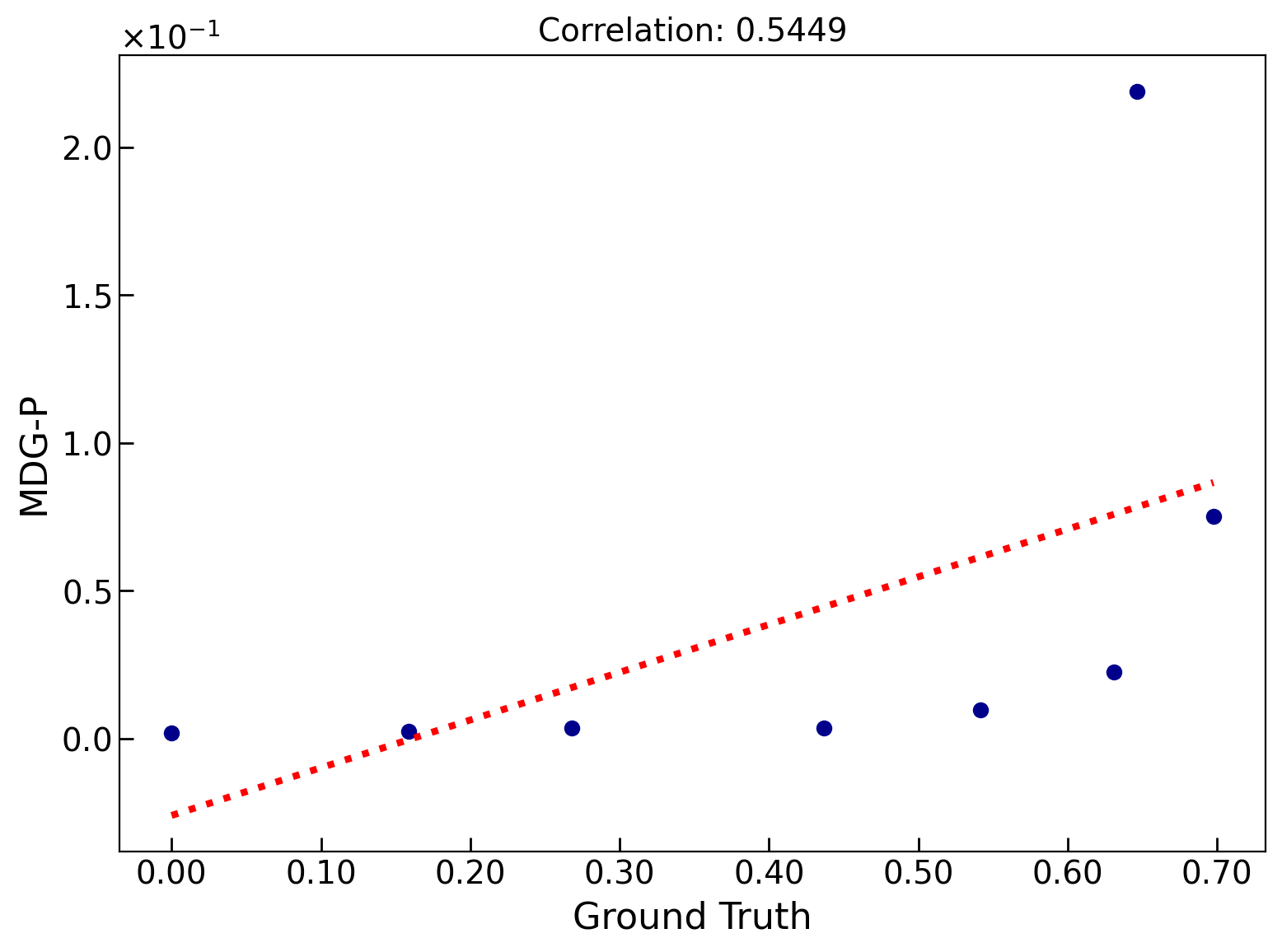}}
\subfloat[S-MIG \cite{hu2022investigating}]{\label{fig6b}
\includegraphics[width=2.2in]{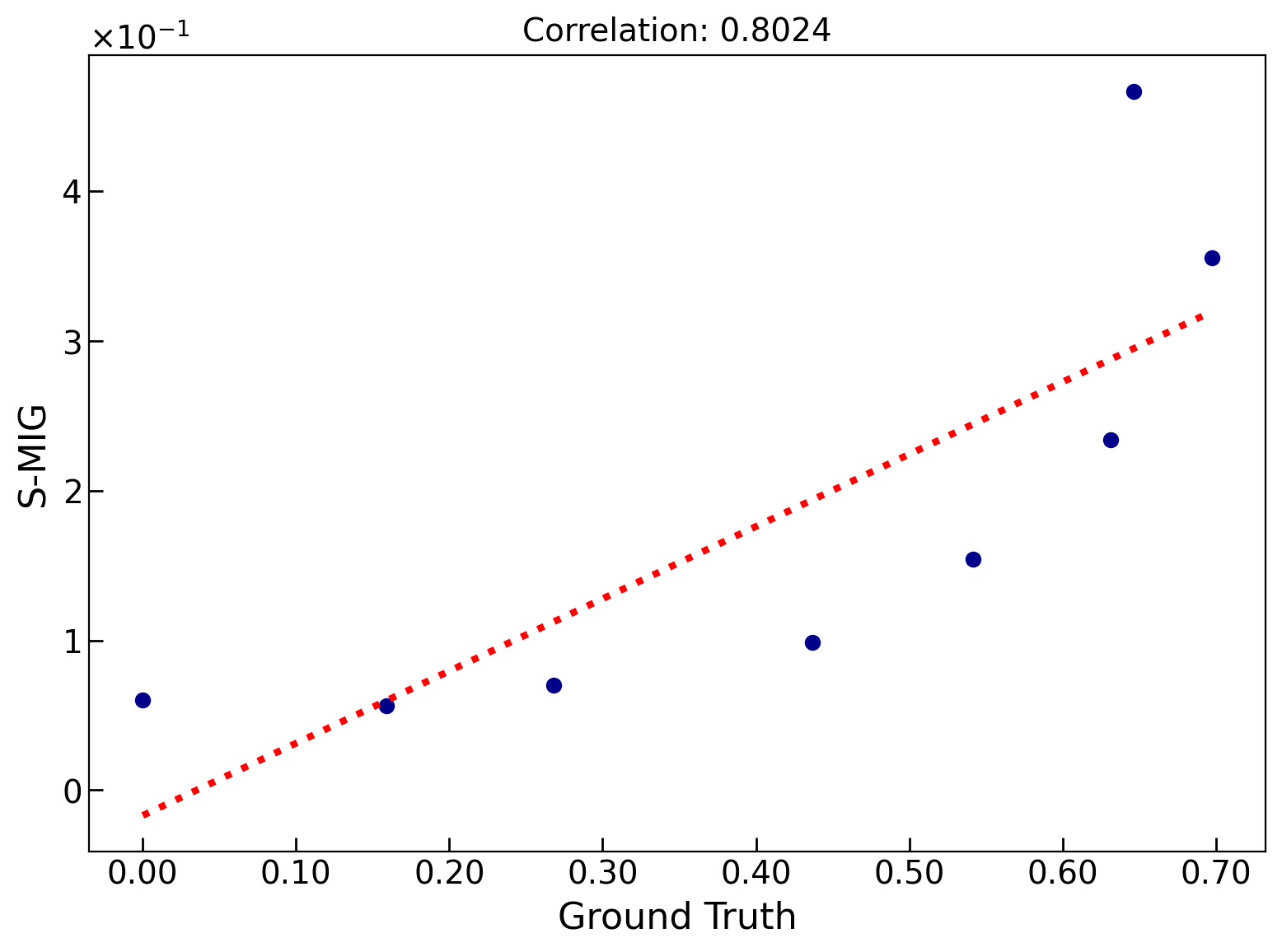}}
\hfill
\subfloat[Perception entropy \cite{roos2021framework}]{\label{fig6c}
\includegraphics[width=2.2in]{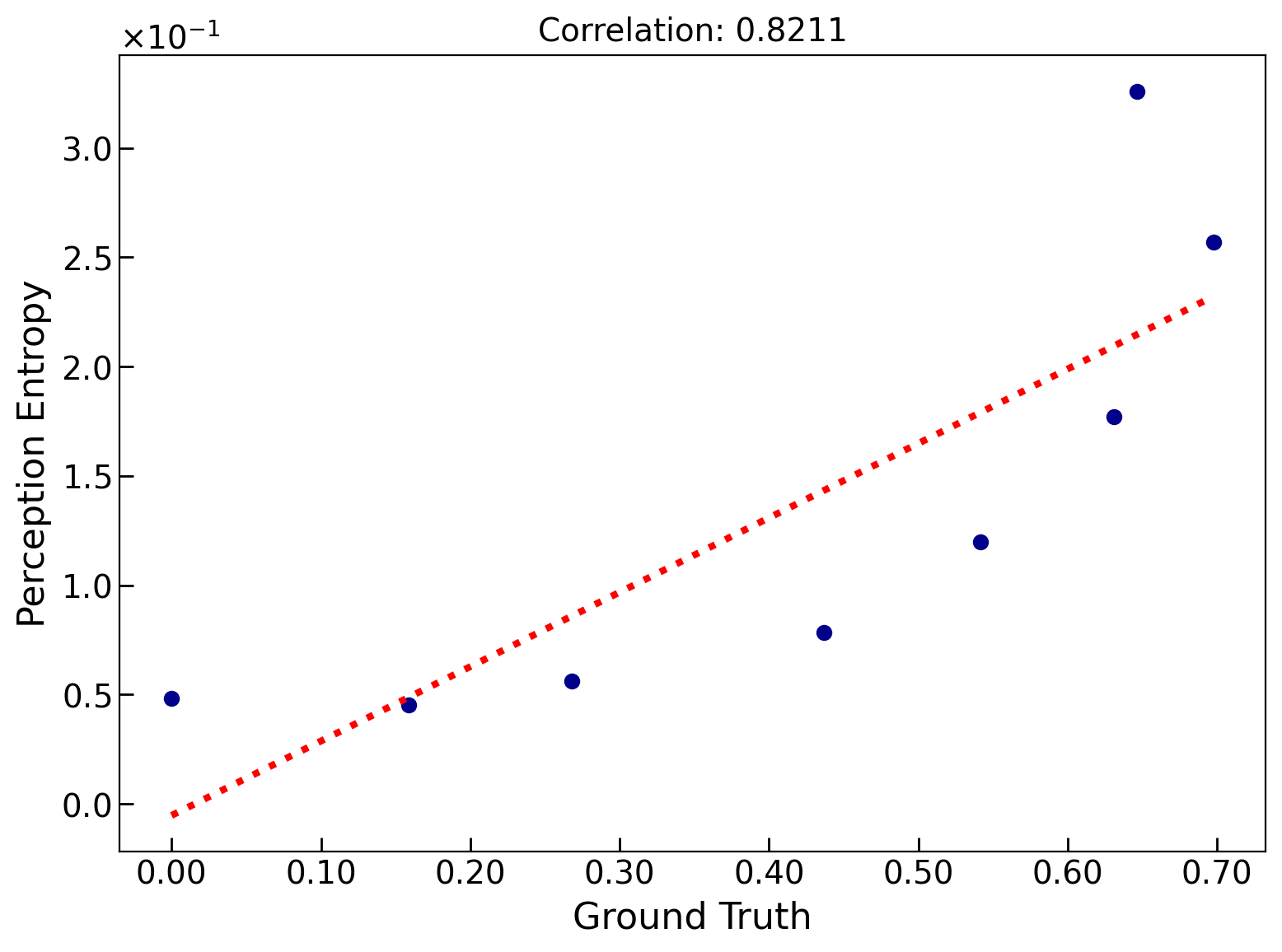}}
\subfloat[PE-VGOP (Our)]{\label{fig6d}
\includegraphics[width=2.2in]{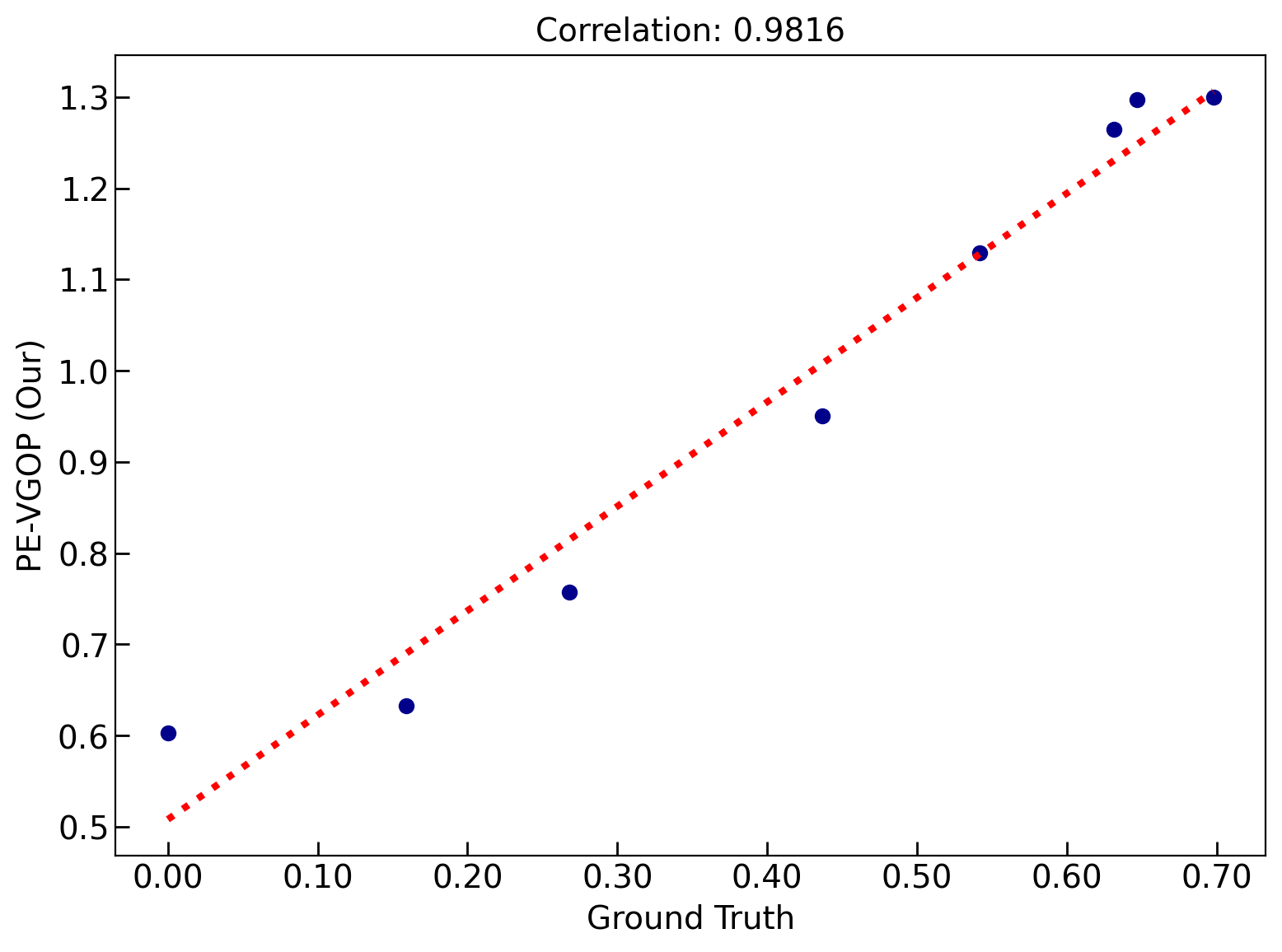}}
\caption{The correlation between alternative metrics and the vehicle detection ground truth in evaluating LiDAR perception performance.}
\label{fig6}
\end{figure*}

As shown in Fig.~\ref{fig6}, we also calculated the correlation between the alternative metrics and the ground truth of vehicle detection performance. From Fig.~\ref{fig6a} to \ref{fig6c}, it is evident that the correlation between the SOTA metrics and vehicle detection ground truth is notably low, with significant outliers. For instance, when the detection performance is around 0.7, the evaluation results of the SOTA metrics show significant variability. The reason for this issue is that the SOTA metrics primarily rely on the quantity or density of point clouds and occupied voxels for evaluation. However, these characteristics cannot accurately reflect vehicle detection performance, as they overlook the impact of point cloud distribution on vehicle detection results. Specifically, vehicle localization and size detection rely more on the distribution characteristics of point clouds rather than on their quantity and density. In contrast, the proposed PE-VGOP metric exhibits a robust correlation with the detection ground truth, with correlation coefficients exceeding 0.98. The above results indicate that the proposed PE-VGOP can accurately and quickly evaluate LiDAR perception performance while demonstrating strong stability.

\section{Case Studies of LiDAR Deployment Optimization}
We optimized the deployments of 16-beam, 32-beam, and 80-beam LiDARs through the proposed optimization model to verify is. The optimization results for the three LiDAR deployments were validated in real-world scenarios and compared with the baseline deployments to assess vehicle detection performance. 

\subsection{Experimental Setup}
In the case studies, the scenario for LiDAR deployment optimization is a five-lane urban road. Given the limitations imposed by the roadside infrastructure and the scenario itself, we focused our optimization efforts solely on the LiDAR placement height and tilt angle along the X-axis. Additionally, we utilized vehicle locations from the NGSIM I-80 vehicle trajectory dataset \cite{ngsim} as inputs to the simulator. This choice was made because the case study scenarios closely match the NGSIM dataset in terms of the number of lanes and road geometry, and the dataset provides accurate vehicle locations.

In the optimization process, the resolution of the vehicle grid is set at $\mu_t = \mu_s = \mu_f = 0.05\text{m}\times0.05\text{m}$; the threshold of VGOP $\delta = 0.005$; the constant loss $C=-1$; the iteration number $T=100$; the particle swarm number $N=20$; the inertia weight $\omega_1=0.7$ and the differential weight $\omega_2=0.5$; the acceleration factors $a_1=0.3$ and $a_2=0.2$; and the differential threshold $\gamma=0.1$. The optimized LiDAR models and parameters are listed in Table \ref{table-4-1}. Considering the limitations of experimental equipment and operational complexity, the optimization space for LiDAR deployment height ranges from 2 to 4.5 meters, while the optimization space for the tilt angle ranges from 0 to 25 degrees.

\begin{table}[!ht]
\centering
\caption{LiDAR Models Used in Experiments \label{table-4-1}}
\begin{tabular}{cccc}
\toprule
\textbf{LiDAR Models} & \textbf{Beam Distribution} & \textbf{FOV} & \textbf{Range} \\ 
\midrule
RS-16 & \adjustbox{valign=c}{\includegraphics[width=1in]{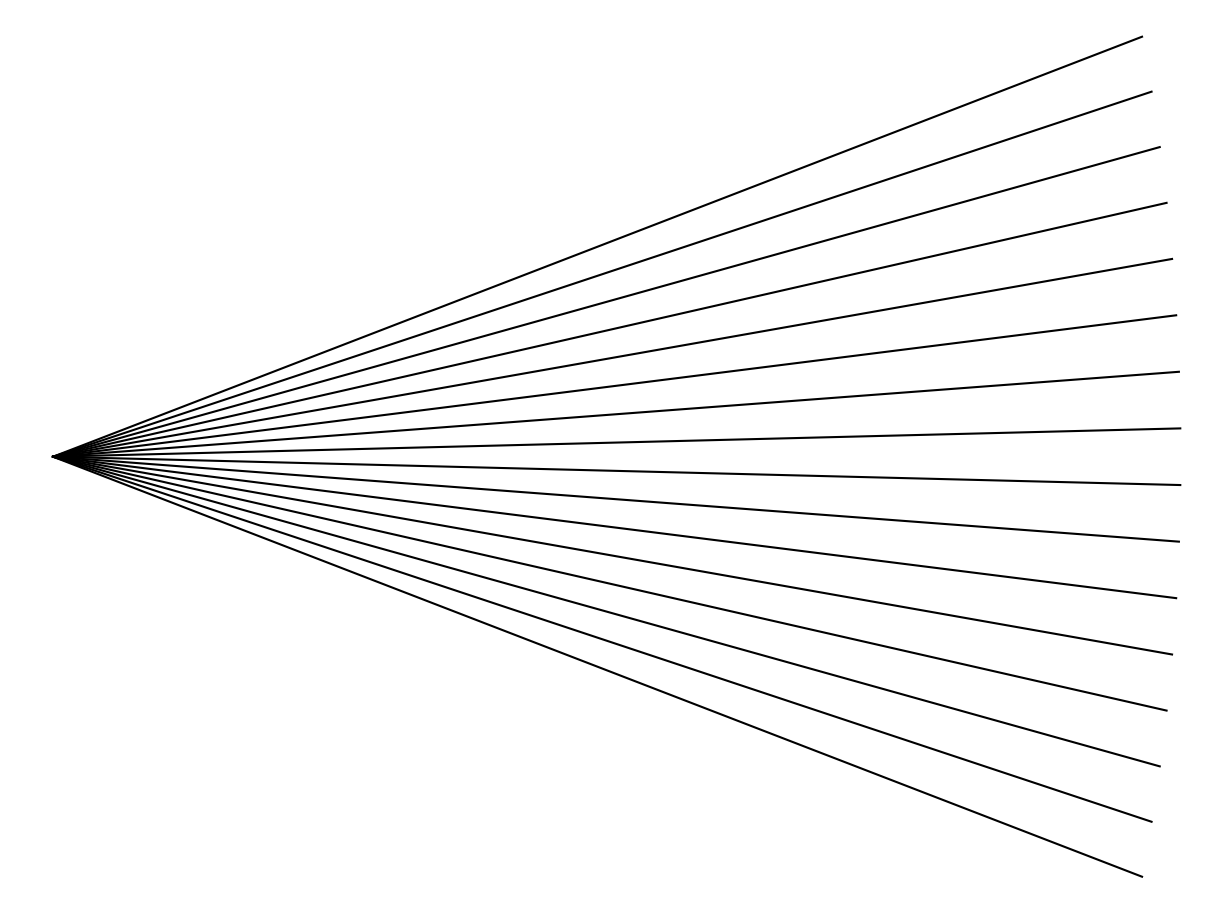}} & $-15^{\circ} \sim +15^{\circ}$ & 50 m \\ 
RS-32 & \adjustbox{valign=c}{\includegraphics[width=1in]{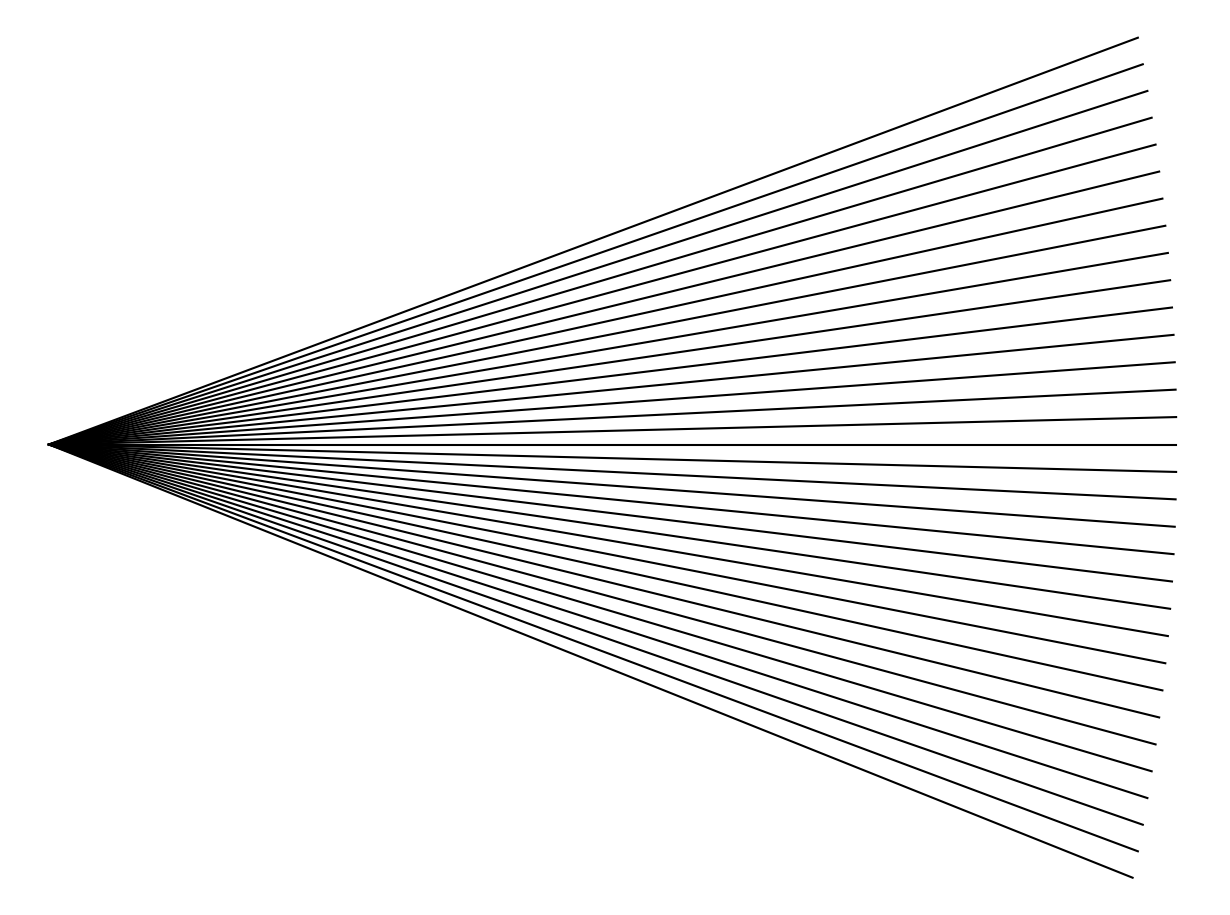}} & $-16^{\circ} \sim +15^{\circ}$ & 80 m \\ 
RS-80 & \adjustbox{valign=c}{\includegraphics[width=1in]{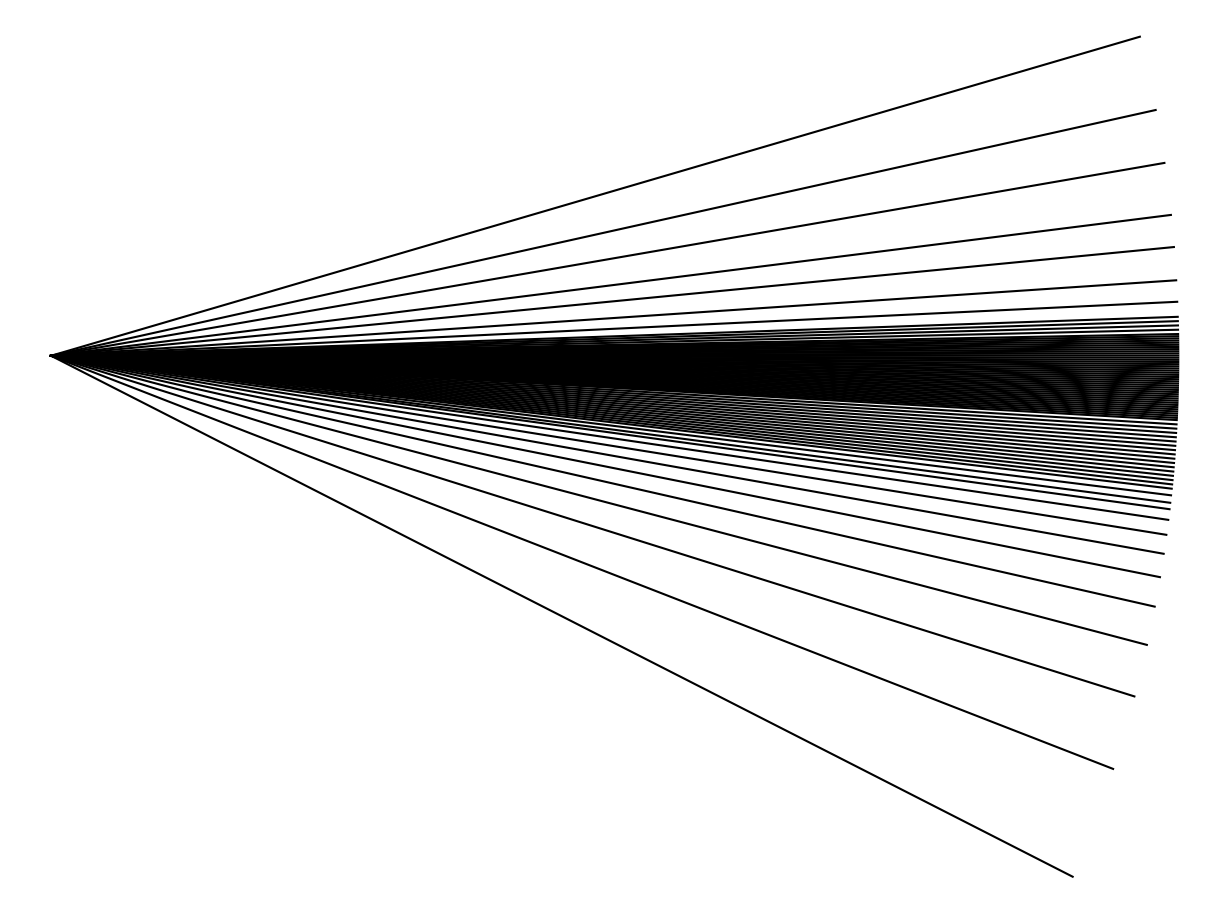}} & $-25^{\circ} \sim +15^{\circ}$ & 150 m \\ 
\bottomrule
\end{tabular}
\end{table}

\begin{figure}[!ht]
\centering
\includegraphics[width=3.45in]{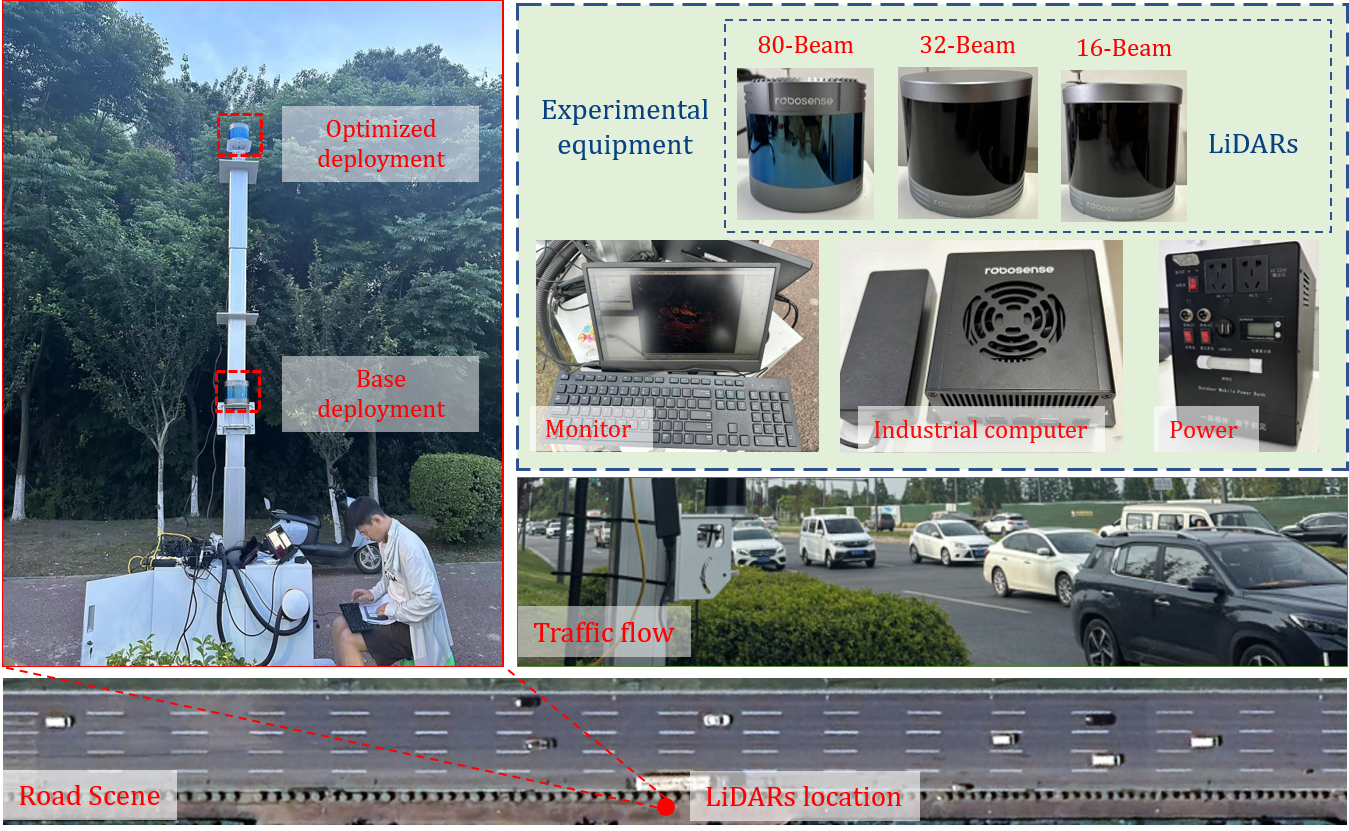}
\caption{The validation experiments in real-world.}
\label{fig7}
\end{figure}

As illustrated in Fig. \ref{fig7}, two LiDAR units were used for each LiDAR model in these experiments, collecting point clouds under both the baseline deployments and optimized deployments. Based on the recommendations from previous studies  \cite{WU2018238,wu2020automatic,zhang2019automatic}, an installation height of 2.0 meters without a tilt angle was selected as the baseline deployment. LiDAR units were mounted on a gimbal platform through a universal joint, allowing for the adjustment of mounting height and tilt angle to achieve both baseline and optimal deployments. From the collected point cloud data, 1,000 frames were selected for each LiDAR in both the baseline and optimized deployments for subsequent analysis. An open-source annotation tool \cite{li2020sustech} was used to annotate the positions, sizes, and headings of vehicles in the point cloud data. In the selected frames, 9,959, 19,949, and 27,497 vehicle samples were annotated for the 16, 32, and 80-beam LiDARs, respectively. These annotations were used as ground truth to evaluate the perception performance of the LiDARs before and after optimization.

\subsection{Results and Analyses}

The LiDAR deployment optimization results are listed in Table \ref{table-4-2}. It is evident that a greater number of LiDAR beams is more suitable for higher installation heights to achieve a broader field of view. Conversely, with fewer LiDAR beams, the lasers can be focused on the perception area by employing a larger tilt angle. By optimizing the deployment in this manner, we analyzed the results of the optimized LiDAR deployments regarding vehicle point cloud distribution and detection performance.

\begin{table}[!ht]
\centering
\caption{LiDAR Deployment Configuration Optimization Results \label{table-4-2}}
\begin{tabular}{ccc}
\toprule
\textbf{LiDAR Models} & \begin{tabular}[c]{@{}c@{}}\textbf{Deployment Height}\\ (meters)\end{tabular} & \begin{tabular}[c]{@{}c@{}}\textbf{Tilt Angle} \\ (degree) \end{tabular} \\ 
\midrule
RS-16      & 4.1                   & 19         \\ 
RS-32      & 4.2                   & 13         \\ 
RS-80      & 4.4                   & 9          \\ 
\bottomrule
\end{tabular}
\end{table}

\subsubsection{Vehicle Point Cloud Distribution}
\
\par
\begin{figure*}[!ht]
\centering  
\subfloat[\footnotesize RS-16 LiDAR]{\label{fig8a}
\includegraphics[width=2.08in ]{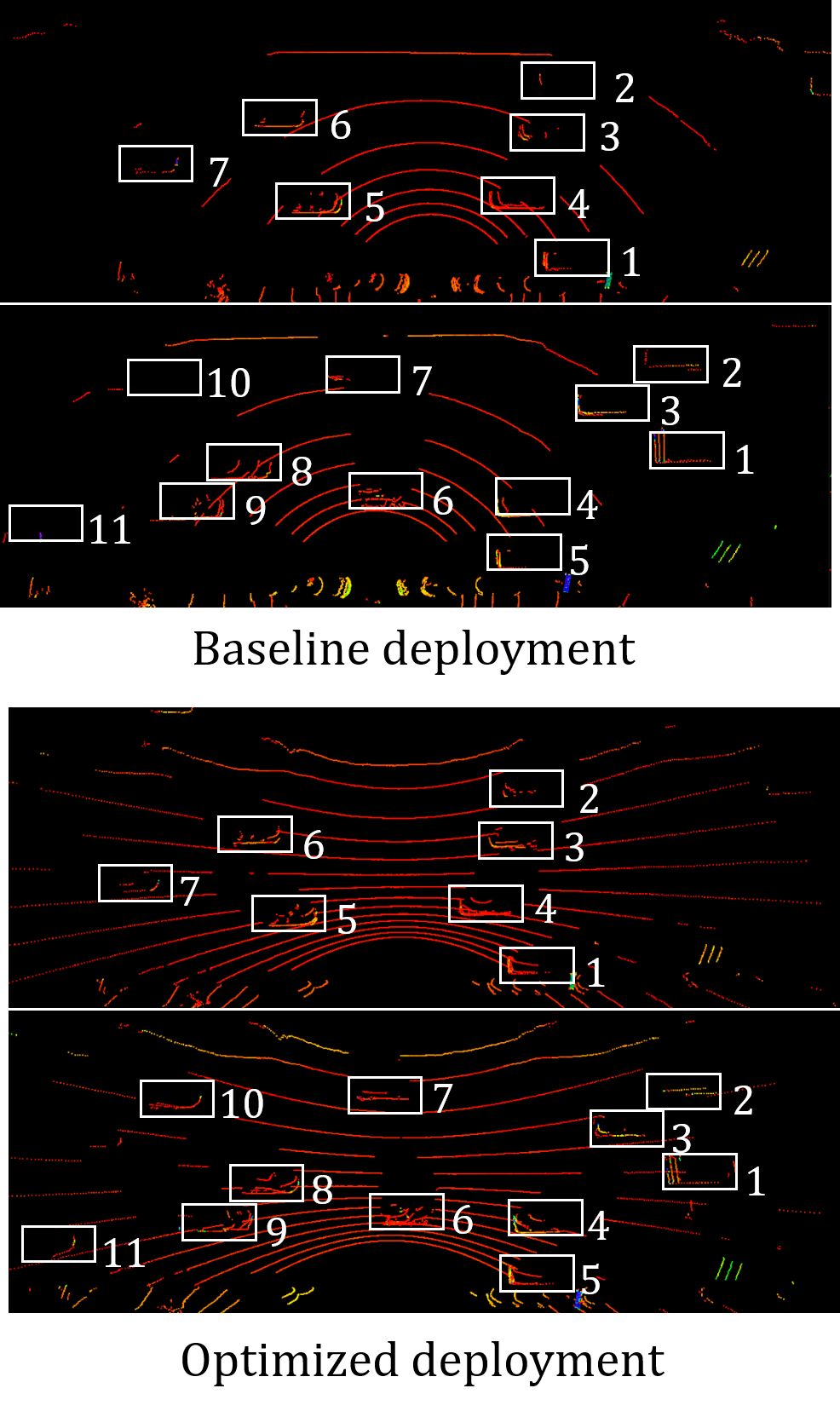}}
\hfil
\subfloat[\footnotesize RS-80 LiDAR]{\label{fig8b}
\includegraphics[width=4.85in]{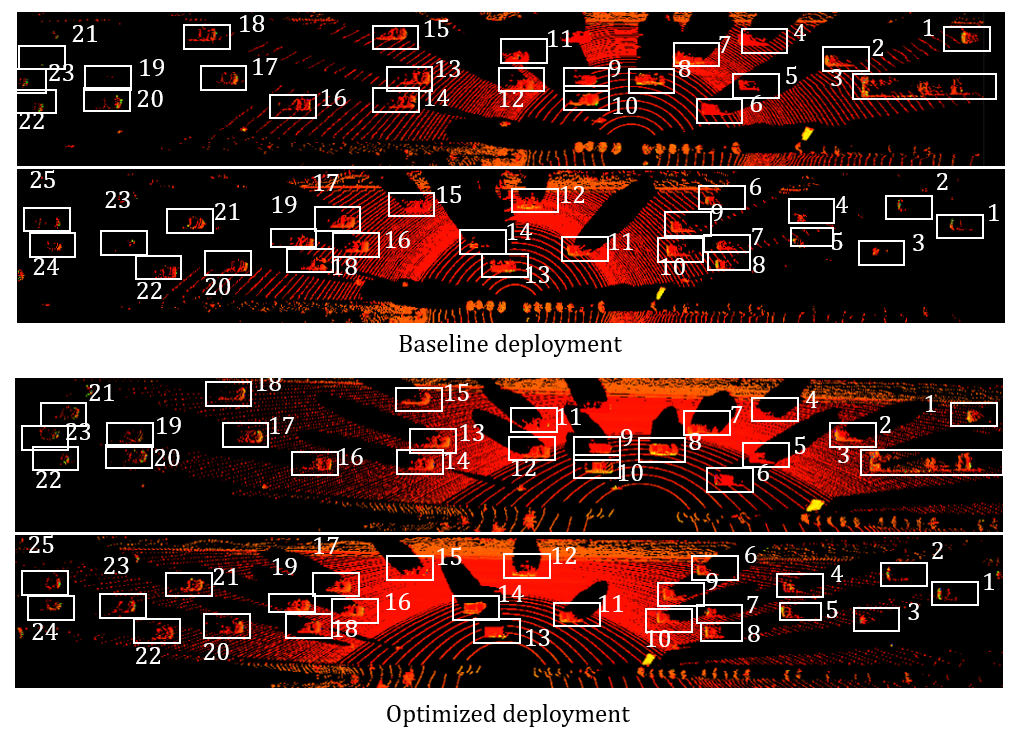} }
\hfil
\subfloat[\footnotesize RS-32 LiDAR]{\label{fig8c}
\includegraphics[width=7in]{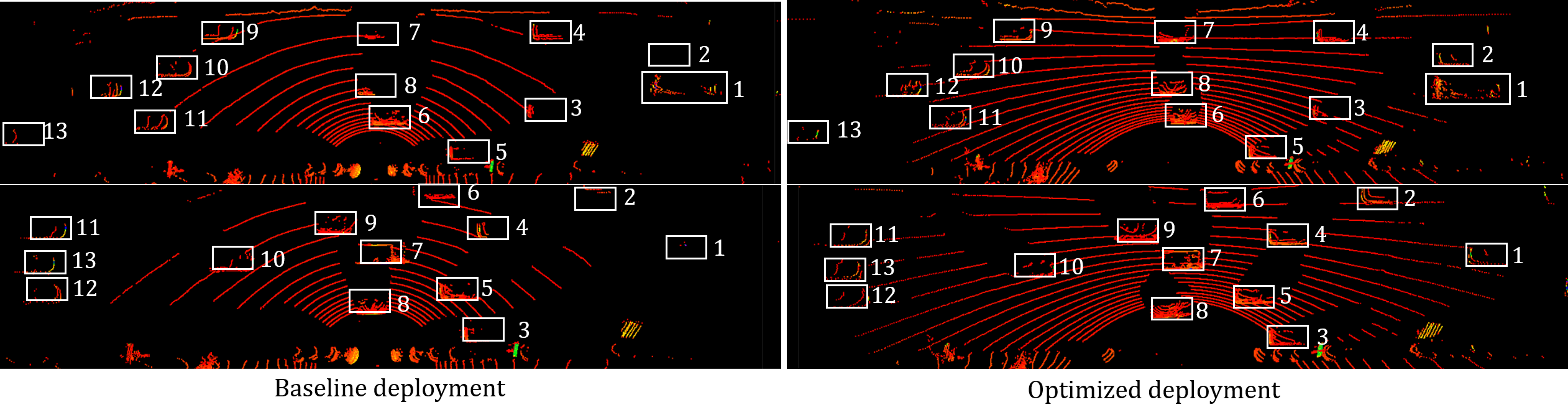}}
\caption{Comparison of point cloud with the baseline LiDAR deployments and the optimized LiDAR deployments.}
\label{fig8}
\end{figure*}

A visualized comparison of point clouds obtained from LiDARs under the baseline deployments and optimized deployments is shown in Fig. \ref{fig8}. It is evident that the proposed LiDAR deployment optimization framework significantly improved the point cloud quantity. On the one hand, the optimizing LiDAR height reduces the shadow areas caused by vehicle obstructions, as shown in Fig. \ref{fig8a}  and Fig. \ref{fig8c}. This reduction enhances LiDAR's ability to detect distant vehicles, such as vehicle No. 11 in Fig. \ref{fig8a} and vehicle No. 2 in Fig. \ref{fig8c}. On the other hand, optimizing the tilt angle allows the laser beams to concentrate within the perception area, increasing the density of the vehicle point clouds (e.g., vehicles No. 20 to 23 in Fig. \ref{fig8b}).

\subsubsection{Vehicle Detection Performance}
\
\par
In this study, vehicle detection performance is evaluated by \textit{Recall}, which reflects the proportion of detected vehicles to the total number of vehicles.
\begin{equation} 
\label{equ-4-2}
Recall = \frac{TP}{TP+FN}  
\end{equation}
where $TP$ (True Positives) represents the number of vehicles that were correctly detected and $FN$ (False Negatives) represents the number of vehicles that were not detected. In this study, \textit{Recall} is obtained by comparing the filtered detection results at various confidence levels with the annotated ground truth rather than by calculating IoU. Because the LiDAR coordinate systems cannot be aligned under baseline and optimized deployments in the experiments, and IoU cannot be calculated correctly.

Table~\ref{table-5-3} presents the vehicle detection \textit{Recall} in baseline deployments and optimized deployments. It is observed that after optimizing the LiDAR deployments, the vehicle detection performance of all three types of LiDAR has improved substantially. When the detection confidence is $\ge$ 0.1, the RS-32 LiDAR showed the most significant improvement of 25\%. The RS-16 and RS-80 LiDARs also show improvements of 8\% to 17\% at different detection confidence levels. These results indicate that the proposed LiDAR deployment optimization model can effectively enhance LiDAR perception capability.

\begin{table*}[!ht]
\centering
\caption{Vehicle detection Recall Comparison for LiDAR Deployment Configurations}
\label{table-5-3}
\begin{tabular}{ccccc}
\toprule
\multirow{2}{*}{\textbf{LiDAR models}} & \multicolumn{2}{c}{\textbf{confidence $\ge$ 0.1}} & \multicolumn{2}{c}{\textbf{confidence $\ge$ 0.3}} \\ \cmidrule(lr){2-3} \cmidrule(lr){4-5} 
                                       & \textbf{baseline deployment}  & \textbf{optimized deployment} & \textbf{baseline deployment} & \textbf{optimized deployment} \\ \midrule
RS-16                                  & 0.70                         & 0.87 (17\%$\uparrow$)            & 0.57                         & 0.67 (10\%$\uparrow$)            \\ 
RS-32                                  & 0.63                         & 0.88 (25\%$\uparrow$)            & 0.53                         & 0.68 (15\%$\uparrow$)            \\ 
RS-80                                  & 0.85                         & 0.95 (10\%$\uparrow$)            & 0.78                         & 0.86 (8\%$\uparrow$)             \\ \bottomrule
\end{tabular}
\end{table*}

\begin{figure*}[!ht]
\centering
\includegraphics[width=5in]{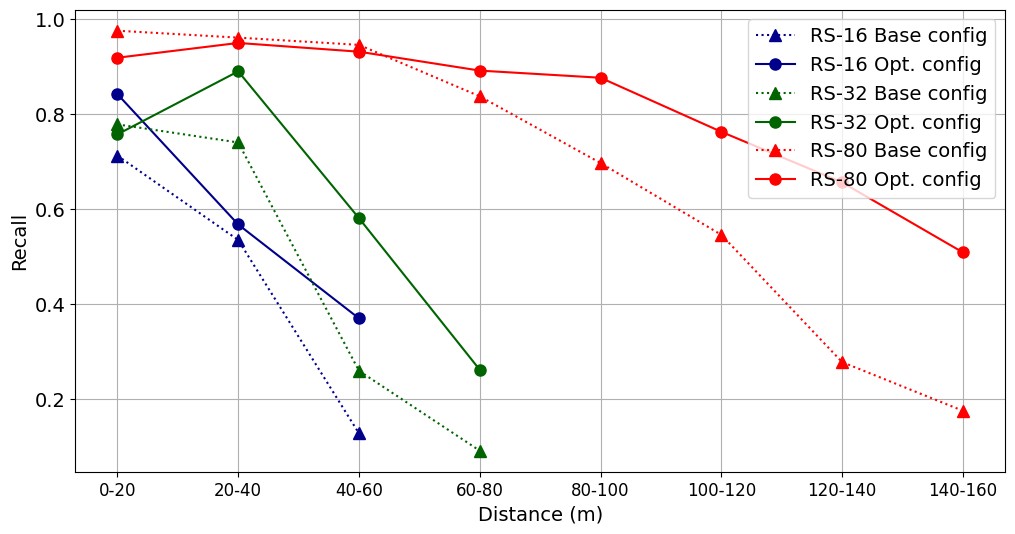}
\caption{Vehicle detection \textit{Recall} comparison for LiDAR deployments at various distance.}
\label{fig9}
\end{figure*}

Additionally, we provided a detailed comparison of the changes in \textit{Recall} with respect to perception distance at detection confidence $\ge$ 0.3, as shown in Fig.~\ref{fig9}. It can be observed that the RS-16 LiDAR shows improved vehicle detection performance across all perception ranges with the optimized deployment configuration. However, the enhancement in perception performance for the RS-32 and RS-80 LiDARs mainly pertains to distant vehicles. For example, compared to the baseline deployment, the optimized deployment of the RS-80 LiDAR resulted in a \textit{Recall} increase of close to 40\% for distances ranging from 120 to 140 meters. However, the RS-80 LiDAR under the baseline deployment configuration performs better in detecting vehicles within 0 to 20 meters. The possible reason for this phenomenon is that the RS-80 LiDAR beams are unevenly distributed within the FOV and fewer beams are allocated for perceiving nearby vehicles (see Table~\ref{table-4-1}). More critically, optimizing the LiDAR deployment through tilt angle cannot resolve this issue. As shown in Fig.~\ref{fig8b}, while the RS-80 LiDAR deployment optimization increases the point cloud distribution for distant vehicles, the point clouds near the LiDAR become sparser. This results in a decrease in detection performance for nearby vehicles.

These results indicate that although optimizing LiDAR deployment can enhance its perception capabilities, it is largely limited by the LiDAR's inherent beam count and distribution. For instance, in most roadside scenarios, LiDAR beams emitted above the horizontal plane are unable to detect targets on the road. Although the tilt angle can adjust the distribution of LiDAR beams, it can also potentially exacerbate issues related to decreased perceptual ability due to uneven distribution, as shown in Fig. \ref{fig8b}. Therefore, it is crucial to consider the perception requirements of the subjects during the LiDAR design phase to optimize the number and distribution of beams.

\section{Conclusion}
This study proposed a LiDAR deployment optimization framework to enhance vehicle detection performance. Initially, we developed a simulator based on Gazebo to collect point cloud data under various LiDAR deployments. The simulator allows for customization of LiDAR models and deployments, simulation of traffic scenarios, and generation of vehicle point clouds. Subsequently, the PE-VGOP metric was introduced to evaluate LiDAR perception performance. A comparative experiment conducted on the KITTI dataset demonstrated that PE-VGOP provides a more accurate and faster evaluation of LiDAR perception performance compared to the ground truth of vehicle detection results. Furthermore, we developed a LiDAR deployment optimization model and designed a DE-PSO solution algorithm focused on LiDAR placement and tilt angle, aimed at maximizing vehicle perception entropy. Field experiments using RS-16, RS-32, and RS-80 LiDARs demonstrated the effectiveness of the proposed optimization model. The results indicate that the proposed optimization framework achieved enhancements in vehicle detection Recall from 8\% to 25\% under different detection confidence thresholds.

Additionally, experiments revealed that the optimization effect of LiDAR deployment on vehicle detection performance depends on the inherent beam distribution of the LiDAR. In future work, we plan to utilize the proposed metric to explore LiDAR perception capability from the perspective of beam number and distribution. For example, we aim to develop a quantitative model to analyze the impact of LiDAR beam distribution and the randomness of vehicles in dynamic traffic flow on perception performance. More importantly, from the LiDAR design perspective, we plan to ensure that each laser scan maximizes the number of collected vehicle point clouds.

\section*{Acknowledgments}
This research was financially supported by the National Natural Science Foundation of China under Grant No.52172395, Natural Science Foundation of Sichuan, China No.2022NSFSC0476.


\begin{thebibliography}{10}
\providecommand{\url}[1]{#1}
\csname url@samestyle\endcsname
\providecommand{\newblock}{\relax}
\providecommand{\bibinfo}[2]{#2}
\providecommand{\BIBentrySTDinterwordspacing}{\spaceskip=0pt\relax}
\providecommand{\BIBentryALTinterwordstretchfactor}{4}
\providecommand{\BIBentryALTinterwordspacing}{\spaceskip=\fontdimen2\font plus
\BIBentryALTinterwordstretchfactor\fontdimen3\font minus \fontdimen4\font\relax}
\providecommand{\BIBforeignlanguage}[2]{{%
\expandafter\ifx\csname l@#1\endcsname\relax
\typeout{** WARNING: IEEEtran.bst: No hyphenation pattern has been}%
\typeout{** loaded for the language `#1'. Using the pattern for}%
\typeout{** the default language instead.}%
\else
\language=\csname l@#1\endcsname
\fi
#2}}
\providecommand{\BIBdecl}{\relax}
\BIBdecl

\bibitem{10208208}
C.~Xiang, C.~Feng, X.~Xie, B.~Shi, H.~Lu, Y.~Lv, M.~Yang, and Z.~Niu, ``Multi-sensor fusion and cooperative perception for autonomous driving: A review,'' \emph{IEEE Intelligent Transportation Systems Magazine}, vol.~15, no.~5, pp. 36--58, 2023.

\bibitem{zhou2022leveraging}
S.~Zhou, H.~Xu, G.~Zhang, T.~Ma, and Y.~Yang, ``Leveraging deep convolutional neural networks pre-trained on autonomous driving data for vehicle detection from roadside lidar data,'' \emph{IEEE Transactions on Intelligent Transportation Systems}, vol.~23, no.~11, pp. 22\,367--22\,377, 2022.

\bibitem{wu2020automatic}
J.~Wu, H.~Xu, J.~Zheng, and J.~Zhao, ``Automatic vehicle detection with roadside lidar data under rainy and snowy conditions,'' \emph{IEEE Intelligent Transportation Systems Magazine}, vol.~13, no.~1, pp. 197--209, 2020.

\bibitem{cai2023analyzing}
X.~Cai, W.~Jiang, R.~Xu, W.~Zhao, J.~Ma, S.~Liu, and Y.~Li, ``Analyzing infrastructure lidar placement with realistic lidar simulation library,'' in \emph{2023 IEEE International Conference on Robotics and Automation (ICRA)}.\hskip 1em plus 0.5em minus 0.4em\relax IEEE, 2023, pp. 5581--5587.

\bibitem{xu2023opencda}
R.~Xu, H.~Xiang, X.~Han, X.~Xia, Z.~Meng, C.-J. Chen, C.~Correa-Jullian, and J.~Ma, ``The opencda open-source ecosystem for cooperative driving automation research,'' \emph{IEEE Transactions on Intelligent Vehicles}, vol.~8, no.~4, pp. 2698--2711, 2023.

\bibitem{hu2022investigating}
H.~Hu, Z.~Liu, S.~Chitlangia, A.~Agnihotri, and D.~Zhao, ``Investigating the impact of multi-lidar placement on object detection for autonomous driving,'' in \emph{Proceedings of the IEEE/CVF Conference on Computer Vision and Pattern Recognition}, 2022, pp. 2550--2559.

\bibitem{roos2021framework}
S.~Roos, T.~V{\"o}lkel, J.~Schmidt, L.~Ewecker, and W.~Stork, ``A framework for simulative evaluation and optimization of point cloud-based automotive sensor sets,'' in \emph{2021 IEEE International Intelligent Transportation Systems Conference (ITSC)}.\hskip 1em plus 0.5em minus 0.4em\relax IEEE, 2021, pp. 3231--3237.

\bibitem{xu2021spg}
Q.~Xu, Y.~Zhou, W.~Wang, C.~R. Qi, and D.~Anguelov, ``Spg: Unsupervised domain adaptation for 3d object detection via semantic point generation,'' in \emph{Proceedings of the IEEE/CVF International Conference on Computer Vision}, 2021, pp. 15\,446--15\,456.

\bibitem{chen2017multi}
X.~Chen, H.~Ma, J.~Wan, B.~Li, and T.~Xia, ``Multi-view 3d object detection network for autonomous driving,'' in \emph{Proceedings of the IEEE conference on Computer Vision and Pattern Recognition}, 2017, pp. 1907--1915.

\bibitem{wu2020deep}
Y.~Wu, Y.~Wang, S.~Zhang, and H.~Ogai, ``Deep 3d object detection networks using lidar data: A review,'' \emph{IEEE Sensors Journal}, vol.~21, no.~2, pp. 1152--1171, 2020.

\bibitem{mao20223d}
J.~Mao, S.~Shi, X.~Wang, and H.~Li, ``3d object detection for autonomous driving: A review and new outlooks,'' \emph{arXiv preprint arXiv:2206.09474}, vol.~1, 2022.

\bibitem{lambert2020performance}
J.~Lambert, A.~Carballo, A.~M. Cano, P.~Narksri, D.~Wong, E.~Takeuchi, and K.~Takeda, ``Performance analysis of 10 models of 3d lidars for automated driving,'' \emph{IEEE Access}, vol.~8, pp. 131\,699--131\,722, 2020.

\bibitem{li2024optimizing}
Y.~Li, L.~Kong, H.~Hu, X.~Xu, and X.~Huang, ``Optimizing lidar placements for robust driving perception in adverse conditions,'' \emph{arXiv preprint arXiv:2403.17009}, 2024.

\bibitem{arefkhani2023sensor}
H.~Arefkhani and Y.~Shafahi, ``Sensor location models with reliable optimal solution for the observation of origin--destination matrix and route flows,'' \emph{Journal of Intelligent Transportation Systems}, pp. 1--20, 2023.

\bibitem{10413581}
L.~A. Klein, ``Roadside sensors for traffic management,'' \emph{IEEE Intelligent Transportation Systems Magazine}, pp. 2--26, 2024.

\bibitem{moshiri2017evaluation}
M.~Moshiri and J.~Montufar, ``Evaluation of detection sensitivity and count performance of advanced vehicle detection technologies at a signalized intersection,'' \emph{Journal of Intelligent Transportation Systems}, vol.~21, no.~1, pp. 52--62, 2017.

\bibitem{cao2019adversarial}
Y.~Cao, C.~Xiao, B.~Cyr, Y.~Zhou, W.~Park, S.~Rampazzi, Q.~A. Chen, K.~Fu, and Z.~M. Mao, ``Adversarial sensor attack on lidar-based perception in autonomous driving,'' in \emph{Proceedings of the 2019 ACM SIGSAC conference on computer and communications security}, 2019, pp. 2267--2281.

\bibitem{9055238}
F.~Zhan and B.~Ran, ``Data accuracy oriented method for deploying fixed and mobile traffic sensors along a freeway,'' \emph{IEEE Intelligent Transportation Systems Magazine}, vol.~14, no.~1, pp. 173--186, 2022.

\bibitem{ji2023optimization}
Y.~Ji, Z.~Yang, Z.~Zhou, Y.~Huang, J.~Cao, L.~Xiong, and Z.~Yu, ``Optimization of roadside sensors placement for cooperative vehicle-infrastructure system,'' in \emph{2023 IEEE 26th International Conference on Intelligent Transportation Systems (ITSC)}.\hskip 1em plus 0.5em minus 0.4em\relax IEEE, 2023, pp. 4813--4819.

\bibitem{ma2021perception}
T.~Ma, Z.~Liu, and Y.~Li, ``Perception entropy: A metric for multiple sensors configuration evaluation and design,'' \emph{arXiv preprint arXiv:2104.06615}, 2021.

\bibitem{liu2021survey}
Y.~Liu, P.~Sun, N.~Wergeles, and Y.~Shang, ``A survey and performance evaluation of deep learning methods for small object detection,'' \emph{Expert Systems with Applications}, vol. 172, p. 114602, 2021.

\bibitem{zhou2019iou}
D.~Zhou, J.~Fang, X.~Song, C.~Guan, J.~Yin, Y.~Dai, and R.~Yang, ``Iou loss for 2d/3d object detection,'' in \emph{2019 International Conference on 3D Vision (3DV)}.\hskip 1em plus 0.5em minus 0.4em\relax IEEE, 2019, pp. 85--94.

\bibitem{vijay2021optimal}
R.~Vijay, J.~Cherian, R.~Riah, N.~De~Boer, and A.~Choudhury, ``Optimal placement of roadside infrastructure sensors towards safer autonomous vehicle deployments,'' in \emph{2021 IEEE International Intelligent Transportation Systems Conference (ITSC)}.\hskip 1em plus 0.5em minus 0.4em\relax IEEE, 2021, pp. 2589--2595.

\bibitem{jin2022novel}
S.~Jin, Y.~Gao, F.~Hui, X.~Zhao, C.~Wei, T.~Ma, and W.~Gan, ``A novel information theory-based metric for evaluating roadside lidar placement,'' \emph{IEEE Sensors Journal}, vol.~22, no.~21, pp. 21\,009--21\,023, 2022.

\bibitem{liu2019should}
Z.~Liu, M.~Arief, and D.~Zhao, ``Where should we place lidars on the autonomous vehicle?-an optimal design approach,'' in \emph{2019 International Conference on Robotics and Automation (ICRA)}.\hskip 1em plus 0.5em minus 0.4em\relax IEEE, 2019, pp. 2793--2799.

\bibitem{mou2018optimal}
S.~Mou, Y.~Chang, W.~Wang, and D.~Zhao, ``An optimal lidar configuration approach for self-driving cars,'' \emph{arXiv preprint arXiv:1805.07843}, 2018.

\bibitem{qu2023seip}
A.~Qu, X.~Huang, and D.~Suo, ``Seip: Simulation-based design and evaluation of infrastructure-based collective perception,'' in \emph{2023 IEEE 26th International Conference on Intelligent Transportation Systems (ITSC)}.\hskip 1em plus 0.5em minus 0.4em\relax IEEE, 2023, pp. 3871--3878.

\bibitem{jiang2023optimizing}
W.~Jiang, H.~Xiang, X.~Cai, R.~Xu, J.~Ma, Y.~Li, G.~H. Lee, and S.~Liu, ``Optimizing the placement of roadside lidars for autonomous driving,'' in \emph{Proceedings of the IEEE/CVF International Conference on Computer Vision}, 2023, pp. 18\,381--18\,390.

\bibitem{kim2019placement}
T.-H. Kim and T.-H. Park, ``Placement optimization of multiple lidar sensors for autonomous vehicles,'' \emph{IEEE Transactions on Intelligent Transportation Systems}, vol.~21, no.~5, pp. 2139--2145, 2019.

\bibitem{koenig2004design}
N.~Koenig and A.~Howard, ``Design and use paradigms for gazebo, an open-source multi-robot simulator,'' in \emph{2004 IEEE/RSJ International Conference on Intelligent Robots and Systems (IROS)(IEEE Cat. No. 04CH37566)}, vol.~3.\hskip 1em plus 0.5em minus 0.4em\relax IEEE, 2004, pp. 2149--2154.

\bibitem{Geiger2012CVPR}
\BIBentryALTinterwordspacing
A.~Geiger, P.~Lenz, C.~Stiller, and R.~Urtasun, ``Vision meets robotics: The kitti dataset,'' \emph{The International Journal of Robotics Research}, vol.~32, no.~11, pp. 1231--1237, 2013. [Online]. Available: \url{https://doi.org/10.1177/0278364913491297}
\BIBentrySTDinterwordspacing

\bibitem{lang2019pointpillars}
A.~H. Lang, S.~Vora, H.~Caesar, L.~Zhou, J.~Yang, and O.~Beijbom, ``Pointpillars: Fast encoders for object detection from point clouds,'' in \emph{Proceedings of the IEEE/CVF conference on computer vision and pattern recognition}, 2019, pp. 12\,697--12\,705.

\bibitem{ngsim}
V.~Alexiadis, J.~Colyar, J.~Halkias, R.~Hranac, and G.~McHale, ``The next generation simulation program,'' p.~22, 2004.

\bibitem{WU2018238}
J.~Wu, H.~Xu, Y.~Zheng, and Z.~Tian, ``A novel method of vehicle-pedestrian near-crash identification with roadside lidar data,'' \emph{Accident Analysis \& Prevention}, vol. 121, pp. 238--249, 2018.

\bibitem{zhang2019automatic}
Z.~Zhang, J.~Zheng, H.~Xu, X.~Wang, X.~Fan, and R.~Chen, ``Automatic background construction and object detection based on roadside lidar,'' \emph{IEEE Transactions on Intelligent Transportation Systems}, vol.~21, no.~10, pp. 4086--4097, 2019.

\bibitem{li2020sustech}
E.~Li, S.~Wang, C.~Li, D.~Li, X.~Wu, and Q.~Hao, ``Sustech points: A portable 3d point cloud interactive annotation platform system,'' in \emph{2020 IEEE Intelligent Vehicles Symposium (IV)}.\hskip 1em plus 0.5em minus 0.4em\relax IEEE, 2020, pp. 1108--1115.

\end{thebibliography}

\newpage

\begin{IEEEbiography}[{\includegraphics[width=1in,height=1.25in,clip,keepaspectratio]{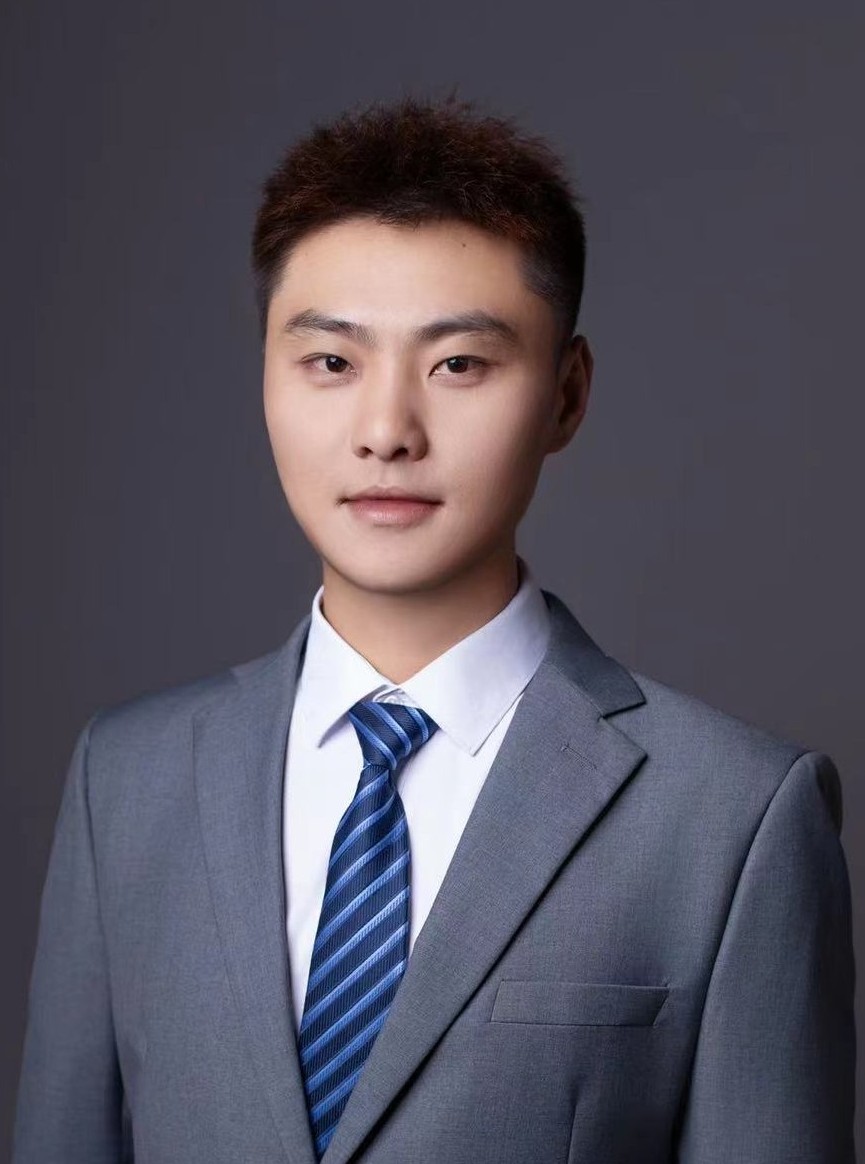}}]{Yongjiang He}
earned his B.S. degree in Transportation Engineering from Shandong University of Science and Technology in 2018 and completed his M.S. degree in the same field at the same institution in 2021. Currently, he is pursuing a Ph.D. in Transportation Engineering at Southwest Jiaotong University. His primary research area is in the field of Intelligent Transportation Systems (ITS), with a focus on sensor-based traffic target detection and sensor deployment optimization. His work aims to contribute to the development of efficient and reliable solutions for enhancing traffic management and safety in urban environments.\end{IEEEbiography}

\begin{IEEEbiography}[{\includegraphics[width=1in,height=1.25in,clip,keepaspectratio]{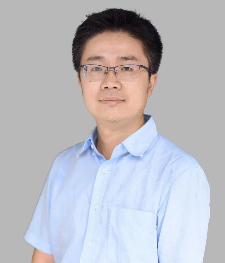}}]{Peng Cao}
earned his B.S. degree in Industrial Engineering from the University of Electronic Science and Technology of China, Chengdu, China in 2009, followed by an M.S. in Management Science and Engineering from Tsinghua University, Beijing, China in 2011, and a Ph.D. in Civil Engineering from Nagoya University, Nagoya, Japan in 2014. He is now an Associate Professor at the School of Transportation and Logistics, Southwest Jiaotong University, Chengdu, China. His research focuses on Traffic Data Collection and Traffic Flow Optimization in the context of Connected Autonomous Vehicles. \end{IEEEbiography}

\begin{IEEEbiography}
[{\includegraphics[width=1in,height=1.25in,clip,keepaspectratio]{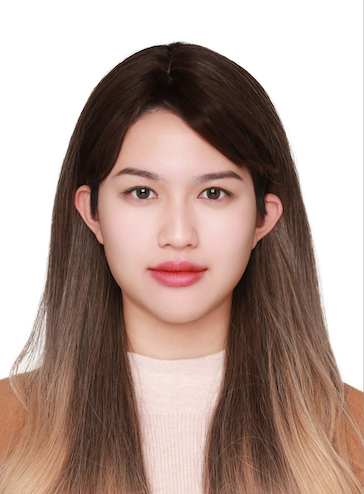}}]{Zhongling Su}
earned her B.S. degree in Traffic Engineering from Southwest Jiaotong University, Chengdu, China in 2020, followed by an M.S. in Traffic Engineering from Southwest Jiaotong University, Chengdu, China in 2023. She is now an Assistant Engineer of High Performance Computing at Shanghai Artificial Intelligence Laboratory, Shanghai, China. Her areas of research include Intelligent Transportation Systems, Traffic Perception, and LiDAR Deployment Optimization.
\end{IEEEbiography}

\begin{IEEEbiography}[{\includegraphics[width=1in,height=1.25in,clip,keepaspectratio]{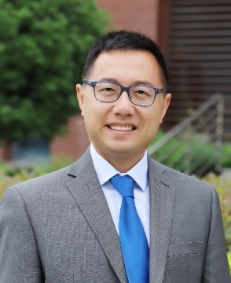}}]{Xiaobo Liu}
earned his Ph.D. degree from the New Jersey Institute of Technology, Newark, New Jersey, in 2004. He is currently a professor with the School of Transportation and Logistics, Southwest Jiaotong University, Chengdu, China. His research focuses on the direction of transportation system analysis under connected vehicle/autonomous vehicle environment, and intelligent logistics analysis. He received the George Krambles Transportation Scholarship, 2003; Most Outstanding Student Paper Award by the Institute of Transportation Engineers Metropolitan Section of New York and New Jersey, 2004; and Stella Dafermos Best Paper Award by the Transportation Research Board Transportation Network Modeling Committee, 2018. He is a committee member on the Standing Committee on Transportation in the Developing Countries (AME40) at the Transportation Research Board. \end{IEEEbiography}
\vfill

\end{document}